%% file: arxiv.tex
\pdfoutput=1
\documentclass[11pt]{article}

\usepackage[]{acl}

\usepackage{times}
\usepackage{latexsym}
\usepackage{amsfonts}

\usepackage[T1]{fontenc}
\usepackage[utf8]{inputenc}

\usepackage{microtype}

\usepackage{inconsolata}

\usepackage{graphicx}

\usepackage{amsmath}
\usepackage{makecell}
\usepackage{lipsum}
\usepackage{enumitem}
\usepackage{graphicx}
\usepackage{booktabs}
\usepackage{multirow}
\usepackage{CJK}
\usepackage{subcaption}
\usepackage{hyperref}
\usepackage{xparse}
\usepackage{pifont}
\usepackage[capitalize]{cleveref}
\usepackage{colortbl}
\usepackage{xcolor}
\definecolor{aliceblue}{RGB}{178, 217, 245}

\definecolor{babyblue}{RGB}{217, 239, 251}

\definecolor{babypink}{RGB}{251, 231, 230}
\definecolor{mygreen}{HTML}{3cb44b}
\definecolor{forestgreen}{RGB}{60, 179, 113}
\definecolor{forestpink}{RGB}{221, 160, 221}
\usepackage{tcolorbox}
\usepackage[linesnumbered,ruled,vlined]{algorithm2e}
\SetKwComment{Comment}{\color{green!50!black}\# }{}

\SetKwProg{Function}{def}{:}{}

\SetKwProg{For}{for}{:}{}
\SetKwProg{If}{if}{:}{}

\newcommand{\ModelName}{\textsc{PRESTO}}
\newcommand{\PipeName}{\textsc{PRESTO}}
\NewDocumentCommand{\daniel}
{ mO{} }{\textcolor{orange}{\textsuperscript{\textit{Daniel}}\textsf{\textbf{\small[#1]}}}}

\NewDocumentCommand{\yuan}
{ mO{} }{\textcolor{purple}{#1}}

\input{math_commands}

\title{\PipeName{}: Progressive Pretraining Enhances Synthetic Chemistry Outcomes}

\author{
He Cao$^{1,2}$\thanks{Equal contribution.}\thanks{Work done during an internship at IDEA.}\quad Yanjun Shao$^{3}$\footnote[1]{}\quad Zhiyuan Liu$^{4}$\quad Zijing Liu$^{1}$ \\ \textbf{Xiangru Tang}$^{3}$\quad  \textbf{Yuan Yao}$^{2}$\quad \textbf{Yu Li}$^{1}$\thanks{Corresponding Author.}\\
$^{1}$International Digital Economy Academy (IDEA)~~~~~~\\$^{2}$Hong Kong University of Science and Technology~~~~~\\ $^{3}$Yale University~~~~~ $^{4}$National University of Singapore\\
\normalsize{\texttt{hcaoaf@connect.ust.hk, yanjun.shao@yale.edu}}
}

\begin{document}
\maketitle

% \twocolumn[{%
%     \renewcommand\twocolumn[1][]{#1}%
%     \vspace{-1cm}
%     \maketitle
%     \begin{center}
%         \vspace{-1.5cm}
%         \includegraphics[width=0.9\textwidth]{figs/teaser.png}
%         % \vspace{0.05cm}
%         \captionof{figure}{\small teaser (place holder)}
%         \label{fig:teaser}
%     \end{center}%
% }]

\input{sec/0-abstract}

\input{sec/1-intro-new-v2}

\input{sec/2-related}

\input{sec/3-method}

\input{sec/3.5-setting}

\input{sec/4-exp}

\input{sec/5-conclusion}

\bibliography{custom}

\appendix
\label{appendix}
\input{sec/suppl}

\end{document}

%% file: math_commands.tex
\usepackage{amsmath,amsfonts,bm}

\def\eqref#1{equation~\ref{#1}}
\def\1{\bm{1}}

\DeclareMathAlphabet{\mathsfit}{\encodingdefault}{\sfdefault}{m}{sl}
\SetMathAlphabet{\mathsfit}{bold}{\encodingdefault}{\sfdefault}{bx}{n}

\definecolor{ForestGreen}{rgb}{0.56, 0.93, 0.56}
\definecolor{SkyBlue}{rgb}{0.53, 0.81, 0.92}
\definecolor{Gold}{rgb}{1.0, 0.84, 0.0}
\definecolor{PeachPuff}{rgb}{1.0, 0.85, 0.73}
\definecolor{Goldenrod}{rgb}{0.85, 0.65, 0.13}
\definecolor{LightGrey}{rgb}{0.83, 0.83, 0.83}
\definecolor{Crimson}{rgb}{0.86, 0.08, 0.24}
\definecolor{LightPink}{rgb}{1.0, 0.71, 0.76}

\newcommand{\adjustedcolorbox}[2]{%
  \begingroup
  \setlength{\fboxsep}{0pt}% 
  \colorbox{#1}{\strut #2}%
  \endgroup
}

\newcommand{\greenbox}[1]{\adjustedcolorbox{ForestGreen}{$\displaystyle #1$}}
\newcommand{\bluebox}[1]{\adjustedcolorbox{SkyBlue}{$\displaystyle #1$}}

\newcommand{\orangebox}[1]{\adjustedcolorbox{PeachPuff}{$\displaystyle #1$}}

\newcommand{\pinkbox}[1]{\adjustedcolorbox{LightPink}{$\displaystyle #1$}}

\newcommand{\bmX}{\bm{X}}

%% file: sec/0-abstract.tex
\begin{abstract}
Multimodal Large Language Models (MLLMs) have seen growing adoption across various scientific disciplines. These advancements encourage the investigation of molecule-text modeling within synthetic chemistry, a field dedicated to designing and conducting chemical reactions to synthesize new compounds with desired properties and applications. Current approaches, however, often neglect the critical role of multiple molecule graph interaction in understanding chemical reactions, leading to suboptimal performance in synthetic chemistry tasks. This study introduces \textbf{\PipeName{}} (\underline{Pr}ogressive Pretraining \underline{E}nhances \underline{S}yn\underline{t}hetic Chemistry \underline{O}utcomes), a new framework that bridges the molecule-text modality gap by integrating a comprehensive benchmark of pretraining strategies and dataset configurations. It progressively improves multimodal LLMs through cross-modal alignment and multi-graph understanding. Our extensive experiments demonstrate that \PipeName{} offers competitive results in downstream synthetic chemistry tasks.
The code can be found at \url{https://github.com/IDEA-XL/PRESTO}.
\end{abstract}

\begin{figure*}[ht]
    \centering
    \includegraphics[width=1.0\textwidth]{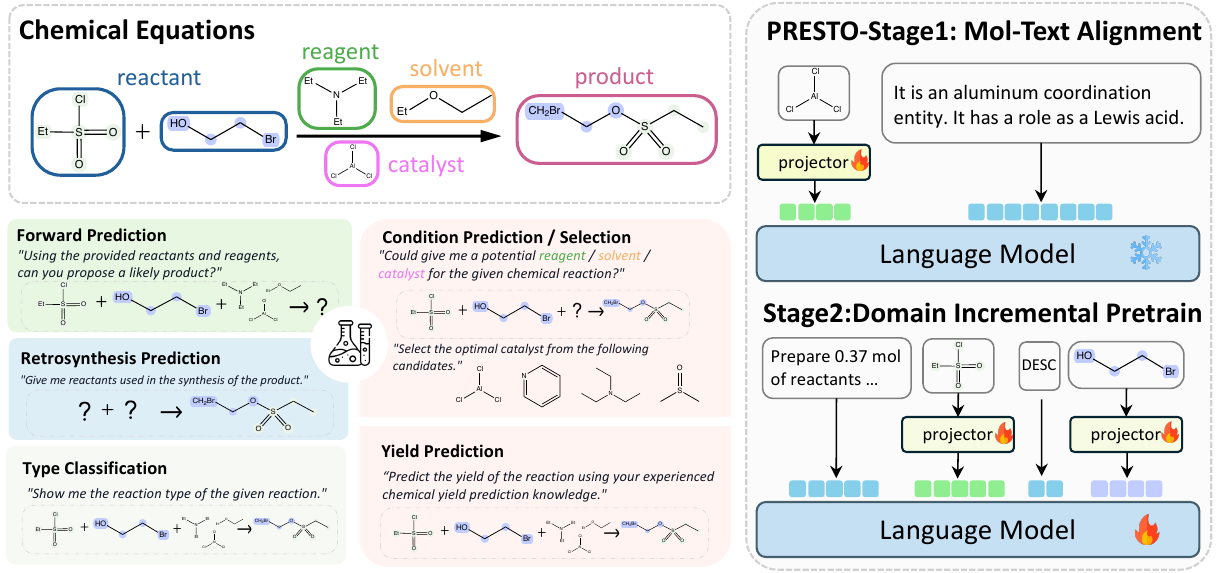} % Replace 'placeholder.png' with your image file
    \caption{\small{Panel~\textbf{(top left)} illustrates the components of a prototypical chemical reaction. Panel~\textbf{(bottom left)} shows the synthetic chemistry tasks that \ModelName{} can support as a dialogue assistant. Panel~\textbf{(right)} provides an overview of the two primary stages in our Progressive Pretraining Strategy \PipeName{}: the Molecule-Text Alignment stage and the Domain Incremental Pretraining stage. These stages enable the evolution from single-graph text modeling to complex interleaved multi-graph text modeling.
    }}
    \label{fig:teaser}
\end{figure*}

%% file: sec/1-intro-new-v2.tex
\section{Introduction}

% para1: multi-modal LM --> molecule-text modeling 
Multi-modal Large Language Models (MLLMs) have achieved extensive success across diverse scientific domains, including medicine \citep{Med-PaLM2}, material science \citep{Jablonka202314EO}, and biochemistry~\cite{LLaVA, LLaVA1.5, BLIP-2}. Motivated by these advances, molecule-text modeling emerges as a new research direction, aiming to bridge the modality gap between molecules and texts~\cite{MoleculeSTM, MolT5}. These methods have shown promising results on molecule captioning, retrieval, and de-novo molecule design~\cite{Git-mol, Text2Mol, 3DMoLM, MolGenSurvey}.

% 有两个贡献，一个是 multi-graph understanding，有两个点：1）pretraining strategy；2）dataset configuration
% 一个是 comprehensive evaluate

% para2: molecule-text modeling --> synthetic chemistry --> disadvantages of prior molecule-text modeling methods, prior single-modal methods 
In this study, we explore molecule-text modeling within synthetic chemistry. Synthetic chemistry involves designing and executing chemical reactions to create new compounds with specific properties and applications. It is a field of immense practical value and includes tasks like forward reaction and retrosynthesis prediction.
Prior molecule-text modeling works~\cite{Mol-Instruction, Text+ChemT5, T5Chem, ChemDFM} have explored synthetic chemistry tasks, but they mostly overlook the 2D molecular graph information. However, 2D molecular graph information is crucial for understanding molecular topologies and is essential for synthetic chemistry in prior graph-based retrosynthesis studies~\cite{GraphRetro, GET}.
% cite several graph-based methods for chemical reaction prediction
On the other hand, while pioneering works~\cite{Git-mol, InstructMol, MolCA, MoMu} have enabled text LLMs to perceive 2D molecular graphs, these methods struggle to process multiple 2D molecular graphs in chemical reactions. This limitation stems from their inadequate exploration and analysis of multi-modal pretraining strategies~\cite{InstructMol, BioMedGPT} and dataset configuration~\cite{DrugChat, 3DMoLM}, which do not fully support the comprehension of multiple graphs:
\begin{itemize}[leftmargin=*]
    \item \textbf{Multi-modal Pretraining Strategy.} The effectiveness of multi-modal LLMs is heavily influenced by their pretraining strategy~\cite{Qwen-VL, VILA, MM1}, involving decisions like tuning or freezing sub-modules at various stages and selecting the granularity of molecular graph representations. The pretraining strategy of existing molecule-text modeling methods varies significantly~\cite{MolCA, MoMu, Git-mol, InstructMol}, creating uncertainty about the most effective approach for synthetic chemistry. Particularly, prior works notably overlook the continual pretraining on synthetic chemistry corpus, which can potentially improve performance. 
    \item \textbf{Dataset Configuration.} The dataset plays a crucial role in the performance of LLMs. For synthetic chemistry tasks, it is evident that including data with multiple molecular graphs in context is essential. However, there is still uncertainty regarding which specific datasets~\cite{PubChem, USPTO_patent, Text2Mol} are most beneficial for synthetic chemistry. Additionally, it remains to be explored whether incorporating single-graph understanding tasks could further enhance performance in synthetic chemistry.
\end{itemize}

% para3: what do we benchmark
To bridge this research gap, we first present a comprehensive benchmark and the corresponding analysis for pretraining strategies and dataset configurations for synthetic chemistry. While several prior benchmarks~\cite{Mol-Instruction, LLaSMol} overlap with synthetic chemistry, they, unfortunately, encompass a limited subset of synthetic chemistry tasks, often mishandle dataset splitting, and sometimes include potential data leakage. We prevent this by cleaning the data meticulously and generating challenging test sets with scaffold splitting. Our analysis shows that progressive multi-modal domain pretraining significantly enhances reaction condition prediction accuracy. Further, we find that increasing the granularity of molecular representation and using interleaved molecule-text data with name-conversion datasets during pretraining improve downstream task performance by better leveraging domain knowledge. 

Building on the insights from our benchmark, we propose \underline{Pr}ogressive Pretraining \underline{E}nhances \underline{S}yn\underline{t}hetic Chemistry \underline{O}utcomes (\PipeName{}), a specialized framework tailored for synthetic chemistry tasks. \PipeName{} enables a MLLM to process and understand interleaved molecular graph-text inputs, deepening the model's grasp of chemical reaction principles by effectively utilizing interactions between molecule-molecule and molecule-text pairs in context. To achieve this, \PipeName{} features a pretraining strategy and a pretraining dataset curated for multi-graph understanding. Specifically, \PipeName{} improves the LLM's performance on synthetic chemistry in two stages progressively: (1) in the first training stage, \PipeName{} cultivates the MLLM's ability of cross-modal alignment; (2) in the second stage, \PipeName{} focuses on multi-graph understanding, and injects domain knowledge of synthetic chemistry into the LLM. Further, to support effective pretraining, we construct a dataset comprising $\sim$3 million samples of synthetic procedure descriptions and molecule name conversions. Through extensive experiments, we demonstrate that \PipeName{} can effectively prepare a multi-modal LLM for downstream tasks of synthetic chemistry. 

% Specifically, we source synthetic procedures from the USPTO-Application corpus~\citep{USPTO_patent}, and identify and convert the molecular entities in text to 2D molecular graphs. Additionally, we include name conversion dataset for both molecular sequences and graphs to deepen the model's understanding of molecule structures.

% Empirical validation confirms that our pretraining dataset allows the model to harness the intrinsic knowledge within chemical synthesis procedures effectively. % 是不是应该挪动到上一段
% Previous approaches \citep{ChemLLM, Mol-Instruction, ChemDFM} mainly focus on constructing text-only SFT datasets, lacking relevance to chemical reaction tasks. We address these issues by integrating all relevant tasks into a multi-objective SFT dataset ($\sim$760k samples), supporting regression, classification, ranking, and generation tasks, spanning over scenarios including product prediction, retrosynthesis, reaction condition recommendation, and yield prediction.

% --------------------------------------------------------------------
% Deep learning has revolutionized synthetic chemistry \citep{Zhou2017OptimizingCR, Liu2017RetrosyntheticRP, Granda2018ControllingAO, Mater2019DeepLI, Ucak2022RetrosyntheticRP, RXNFP} by precisely predicting, optimizing, and elucidating chemical reactions. Using extensive datasets of chemical structures and reaction conditions, it excels in forecasting products, designing molecules, and understanding reactivity principles.

%% file: sec/2-related.tex
\section{Related Works}
% \yuan{[Are the following works relevant to synthetic chemistry, or not? If yes, merge the following paragraph into the first paragraph of related works; If no, merge the following paragraph into the second paragraph of related works.]
% Recently, researchers have turned to the off-the-shelf usage of general-purpose large language models (LLMs) \citep{ChatGPT, GPT4V, claude3} for chemistry tasks \citep{ChemBench, Jablonka2024LeveragingLL}. Despite pretraining on vast general corpora, these models face challenges in maintaining chemical plausibility \citep{ChemLLMBench, PromptEO}, demonstrating limited capabilities compared to task-specific deep learning models.} 

% \yuan{[Need to merge this paragraph with the first paragraph in Related Works.] Prior works have achieved extensive success in synthetic chemistry using LM-based methods~\citep{MolecularTransformer, YieldBERT, RXNFP}. However, these models mostly specialize in a few tasks of synthetic chemistry but struggle to generalize to other synthetic tasks. This is because they are pretrained on domain-specific data, limiting their ability to share knowledge and adapt to new tasks of synthetic chemistry. To resolve the issue, multi-task methods \citep{Text+ChemT5} are explored and show a strong capability of handling tasks across domains. }

\paragraph{Deep Learning for Synthetic Chemistry.}
Synthetic chemistry, a fundamental problem in chemistry, has seen significant advances through deep learning models that assist in various reaction-related tasks using descriptor-based \citep{NeuralSym, NeuralSymNature}, graph-based \citep{GLN, Graph2SMILES}, and sequence-based approaches \citep{MolecularTransformer, Chemformer}. Recent works \citep{T5Chem, RetroSchaller, Mol-Instruction, LLaSMol} also adapt language models for tasks such as forward reaction prediction \citep{MolecularTransformer}, retrosynthesis \citep{Retroformer, ReactXT}, and reaction type classification \citep{RXNFP}, demonstrating high accuracy. Although these models specialize in specific synthetic chemistry tasks, their pretraining on domain-specific data limits their ability to generalize and adapt to other synthetic tasks. To address this issue, multi-task methods \citep{T5Chem, Text+ChemT5} have been explored and demonstrate strong capabilities across domains. However, they are constrained by using only molecular sequences as input, overlooking the potential of textual information to assist in modeling. In contrast, our approach integrates reaction-related textual information with molecular modeling, enabling a flexible adaptation to various downstream tasks.

\paragraph{Molecule \& Text Modeling (MTM).}
The integration of biomolecular modeling with natural language leverages rich textual data sources to enhance understanding and facilitate downstream text-related molecular tasks \citep{MolT5, Text+ChemT5, BioT5, Mol-Instruction, LLaSMol, MolFM}.
Various approaches have been proposed to learn effective representations of molecules, including 1D sequences \citep{MolGen, Chemformer, MolT5, MolecularTransformer, SMILES-BERT}, 2D graphs \citep{GROVER, Graphformer, MolCLR}, 3D conformations \citep{GraphMVP, UniMol} and a combination of them \citep{TransformerM, MolLM}. 
% Cross-modal approaches to fuse molecular representations with text \citep{MoleculeSTM, MoMu, KV-PLM, MolLM} have elicited the ability \daniel{what}, and are the very first attempt to interact with human knowledge. 
Cross-modalities modeling includes contrastive learning over molecules and text \citep{MoMu, MoleculeSTM, MolLM} or unified alignment of the two modalities through language modeling \citep{KV-PLM, GIMLET, MolCA, 3DMoLM}. 
Prior works have primarily focused on individual molecule understanding or molecule-text retrieval, while our research expands to model multiple molecules and contextual text, thereby facilitating tasks relevant to chemical reactions.

% SMILES \citep{SMILES} and SELFIES \citep{SELFIES} are 1D sequences that language models can readily process. With the advent of publicly available language models \citep{Llama, Mistral} and specialized models pretrained on scientific knowledge  \citep{Meditron, Galactica}, recent works \citep{Text+ChemT5, BioT5, Mol-Instruction, LLaSMol} have shown the effectiveness of instruction tuning for molecular tasks, including reaction prediction, property prediction, and molecule generation. Meanwhile, 2D and 3D representations require specialized encoders. \citep{InstructMol, MolCA, 3DMoLM, MolFM} integrate molecular graph representations with the natural language via cross-modal projection in LLMs. These approaches utilize LLMs' multitasking abilities to address various molecular tasks within a unified framework. Their conversational capabilities make them a reliable molecular assistant in drug discovery.

\paragraph{Multi-modal Language Models.}
The multimodal large language models (MLLMs) field has seen rapid progress recently. Several works \citep{Flamingo, GIT, PaLI, InstructBLIP, BLIP-2, KOSMOS-1, LLaVA} have proposed different architectures for integrating visual information into LLMs.
Researchers have explored various strategies for integrating external modalities into LLMs. \citet{VILA} and \citet{MM1} conducted ablation studies on textual and visual data composition during training. \citet{Prisma} examined the design space of MLLMs, including training pipeline, modality representations, and scaling. 
Recent studies have attempted to apply similar methods to small molecule \citep{3DMoLM, InstructMol, DrugChat} or protein domains \citep{InstructProtein, ProtT3}. However, there are very few studies investigating the specific design of training strategies in the biomolecular domain.

%% file: sec/3-method.tex
\section{\PipeName{} Framework}

\subsection{Preliminary}
Here we introduce our model architecture, which follows the common practice in multi-modal LLMs \citep{LLaVA, Qwen-VL, Prisma}.
Formally, our model processes a collection of 2D molecule graphs represented as $\{\greenbox{\bmX_G^{(i)}}\}_{i=1}^n$, along with text prompt tokens $\{\bluebox{\bmX_{T}^{(j)}}\}_{j=1}^m$ describing synthetic processes or task queries. The input sequence is designed to accommodate the interleaved nature of text and molecule tokens, denoted $\{t_k\}_{k=1}^{m+n}$, where each $t_k$ is a text token $\bluebox{\bmX_{T}^{(j)}}$ or a molecule graph $\greenbox{\bmX_G^{(i)}}$. These inputs are processed through 1) a molecular representation encoder, 2) a molecule-language projector, and 3) a language model.

\noindent
\textbf{Molecular Representation.} Each $\greenbox{\bmX_G^{(i)}}$ is first processed by a molecule encoder $f_M$, which outputs a sequence of features $p_M^{(i)}$, such that $p_M^{(i)}=f_M(\greenbox{\bmX_G^{(i)}})$. The length of $p_M^{(i)}$ is variable and depends on the granularity of the representation.

\noindent
\textbf{Molecule-Language Projector.} Next, we map each $p_M^{(i)}$ to embeddings $e_M^{(i)}$ using a learned projector $f_{\psi}$, where $e_M^{(i)} = f_{\psi}(p_M^{(i)})$.

\noindent
\textbf{Language Model.} The interleaved input sequence $\orangebox{\mathcal{E}_I}$ is formed by the ordered union of molecule embeddings $\greenbox{\mathcal{E}_M} = \{e_M^{(i)}\}_{i=1}^n$ and text token embeddings $\bluebox{\mathcal{E}_T} = \{e_T^{(j)}|e_T^{(j)} = f_{\text{embed}}(\bluebox{\bmX_T^{(j)}})\}_{j=1}^m$:
\[
\orangebox{\mathcal{E}_I} = \greenbox{\mathcal{E}_M} \cup_{o} \bluebox{\mathcal{E}_T},
\]
where $\cup_{o}$ preserves the order of elements as they appear in the original input sequence $\{t_k\}_{k=1}^{m+n}$. This interleaved sequence is passed to the language model to generate the output text $\pinkbox{\bmX_O} = \text{LM}_{\theta}(\orangebox{\mathcal{E}_I})$.

\subsection{Training Procedure}
\label{sec:3.2}
Our complete training procedure includes the \PipeName{}'s two-stage pretraining and the downstream supervised finetuning.
% We adhere to the steps: aligning different modalities, constructing domain knowledge from interleaved corpus and fundamental instance recognition tasks, and finally efficiently adapting to downstream tasks.
\noindent
\paragraph{\PipeName{}-Stage1: Molecule-Text Alignment.}
This stage aims to bridge the modality gap between the molecular and textual representations. We start from a pretrained molecule encoder $f_M$, a language model $\text{LM}_{\theta}$, and a randomly initialized molecule-language projector $f_{\psi}$. $f_{\psi}$ is then trained on molecule-text pairs from \citep{PubChem} while freezing the weights of $f_M$ and $\text{LM}_{\theta}$. The template for captioning can be found in Appendix \ref{app:template_pretrain}.
% text data from \citep{PubChem} with $f_M$ and $\text{LM}_{\theta}$ frozen.

\noindent
\paragraph{\PipeName{}-Stage2: Domain Incremental Pretraining.}
During this stage, we continue to train the model on a large corpus of molecule-text pairs with interleaved segments \citep{USPTO_patent, PubChem}. Training on mixed data helps the model further understand the relationships between molecular graphs and text. Both $f_M$ and $\text{LM}_{\theta}$ are updated in this stage. See Appendix \ref{app:template_pretrain} for details of the instruction template.

\noindent
\paragraph{Supervised Fine-Tuning (SFT).} The final stage adapts the pretrained model to a diverse set of downstream tasks by instruction tuning. Similar to \citep{InstructMol, MolCA}, each example consists of input molecules or reactions $\{\greenbox{\bmX_G^{(i)}}\}_{i=1}^n$, a natural language instruction $\{\bluebox{\bmX_{T}^{(j)}}\}_{j=1}^m$, and the target output $\pinkbox{\bmX_O}$. Details of the instruction template can be found in the Appendix \ref{app:template_downstream}.

%% file: sec/3.5-setting.tex
\begin{figure*}[h]
    \centering
    \includegraphics[width=0.95\textwidth]{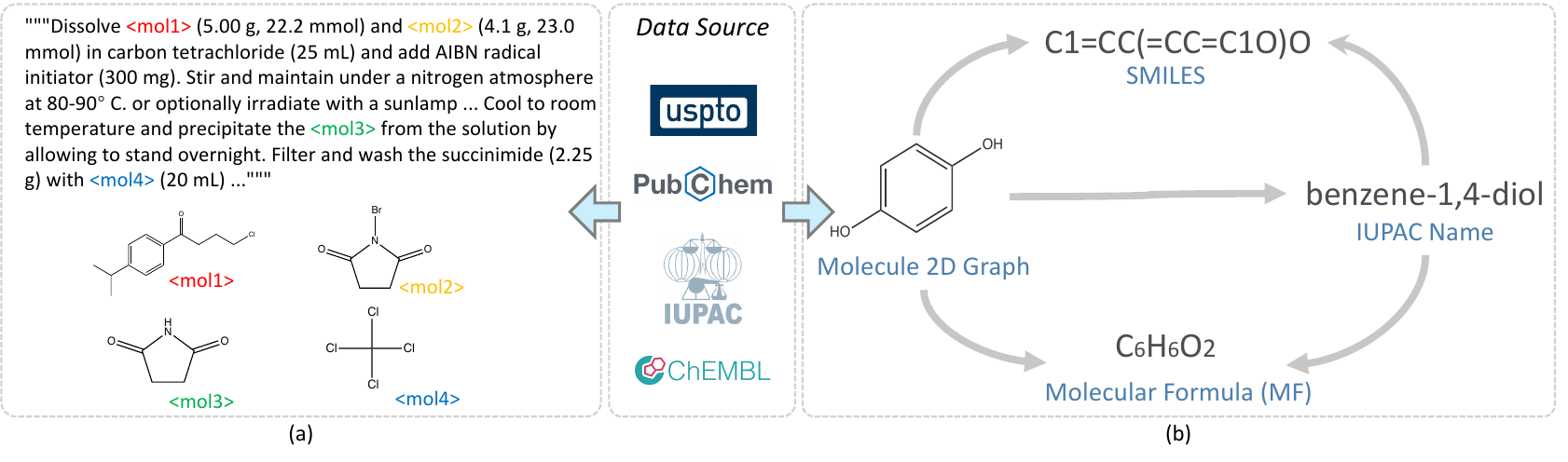} % Replace 'placeholder.png' with your image file
    \vskip -0.2in
    \caption{\small{Panel~\textbf{(a)} illustrates the interleaved molecule-text dataset format, primarily derived from USPTO-Application~\cite{USPTO_patent}. Panel~\textbf{(b)} displays the five tasks included in the Molecular Name Conversion Tasks (directions drawn as arrows), with data mainly sourced from PubChem~\cite{PubChem}, IUPAC~\cite{IUPAC}, and ChEMBL~\cite{ChEMBL}.}}
    \label{fig:example}
    \vskip -0.1in
\end{figure*}

\subsection{Pretrain Dataset}
We present datasets utilized in the \PipeName{} pretraining pipeline. For the first stage of alignment, we use a caption dataset, and for the second stage of domain incremental pretraining, we use an interleaved molecule-text and name-conversion dataset.
\begin{table}[!hbt]
\centering
\small
\setlength{\tabcolsep}{1.6mm}{
\scalebox{0.78}{
\begin{tabular}{lrrrr}
\toprule
\textsc{Task} & \textsc{\# Train} & \textsc{\# Valid} & \textsc{\# Test} & \textsc{\# All} \\
\midrule[1.1pt]
\rowcolor[RGB]{234, 238, 234}
\multicolumn{5}{l}{\textit{Pretrain Stage1: Molecule Caption}} \\
\rowcolor[RGB]{220, 230, 242}
\multicolumn{5}{l}{\textsc{Data Source:} \citet{PubChem}} \\
PubChem Caption & 326,675 & - & - & 326,675 \\
% Interleaved
\midrule[1.1pt]
\rowcolor[RGB]{234, 238, 234}
\multicolumn{5}{l}{\textit{Pretrain Stage2: Interleaved Molecule-Text}} \\
\rowcolor[RGB]{220, 230, 242}
\multicolumn{5}{l}{\textsc{Data Source:} \citet{USPTO_patent}} \\
USPTO-Application & 1,588,709 & - & - & 1,588,709 \\
% Name Conversion
\midrule[1.1pt]
\rowcolor[RGB]{234, 238, 234}
\multicolumn{5}{l}{\textit{Pretrain Stage2: Name Conversion}} \\
\rowcolor[RGB]{220, 230, 242}
\multicolumn{5}{l}{\textsc{Data Source:} \citet{PubChem, LLaSMol}} \\
IUPAC to Formula & 300,000 & 1,497 & 2,993 & 304,490 \\
IUPAC to SMILES & 300,000 & 1,497 & 2,993 & 304,490 \\
Molecule Graph to Formula & 300,000 & 1,497 & 2,993 & 304,490 \\
Molecule Graph to IUPAC & 300,000 & 1,497 & 2,993 & 304,490 \\
Molecule Graph to SMILES & 293,288 & - & - & 293,288 \\
% % Overall
% \midrule[1.1pt]
% \textbf{Overall} & 2,331,344 & 70,013 & 67,669 & 2,469,026 \\
\bottomrule
\end{tabular}
}}
\vskip -0.1in
\caption{\PipeName{} progressive pretraining dataset.}
\vskip -0.2in
\label{tab:dataset_pretrain}
\end{table}

\noindent
\paragraph{Caption Dataset.} We use molecule-text pairs sourced from PubChem \citep{PubChem} for aligning molecule and text modalities. Each molecule structure is associated with a textual description of chemical and physical properties or high-level bioactivity information.

\noindent
\paragraph{Interleaved Molecule-Text Dataset.} 
We start by extracting raw descriptions of experimental procedures from the chemical reaction database USPTO-Applications \citep{USPTO_patent}. 
Further, we use BERN2 \citep{BERN2} to identify all molecule entities in the texts and convert them into 2D molecular graphs. 
% Using BERN2 \citep{BERN2} for entity extraction, we identify all molecule entities in the texts and convert them into 2D molecular graphs. 
We then preprocess the data to remove samples with too many molecule entities or molecules with excessive atom counts to control input length. The resulting interleaved dataset comprises approximately 1.6M samples, covering more than 342K unique molecules. Refer to Appendix~\ref{app:data_interleaved} for detailed processing steps and data statistics.

\noindent
% wiki link to IUPAC / Chemical Formula
\paragraph{Name Conversion Dataset.} 
% This category includes five tasks that involve converting between different molecule representations: (1) IUPAC Names \citep{IUPAC}, (2) Chemical Formula \citep{ChemicalFormula}, (3) SMILES \citep{SMILES}, (4) Molecular Graph. The name conversion tasks serve to further align the graph representation of molecules with their sequential or knowledge representations, providing the model with a basic understanding of the molecular structure and enabling instance recognition of specific molecules. The molecule entities are collected from the PubChem database \citep{PubChem}.
A molecule can be represented by 2D molecular graphs and different 1D sequential representations:  IUPAC names \citep{IUPAC}, chemical formulas \citep{ChemicalFormula}, and SMILES \citep{SMILES}. These 1D sequential representations are used interchangeably in the textual corpus, and each corresponds to a particular aspect of molecular structures. For example, the IUPAC name highlights the subgraph composition of molecules, while SMILES explicitly lists all atom types. Therefore, learning the conversion between these 1D representations and 2D graphs helps the LLM to align different molecular mentions in texts and improves its understanding of molecular structures.

\subsection{Downstream Tasks}
\label{sec:downstream_tasks}
We evaluate \PipeName{} on a diverse set of downstream tasks in synthetic chemistry, as detailed in Table \ref{tab:dataset_downstream}. Our assessment provides a more comprehensive and representative evaluation of downstream tasks, extending beyond the scope of previous benchmarks. The detailed data preprocessing pipeline is provided in the Appendix \ref{app:data_downstream}.

\noindent
\paragraph{Reaction Prediction.} This category includes two tasks: \textit{Forward Prediction}, which involves predicting the product molecules given the reactant molecules, and \textit{Retrosynthesis}, which predicts the reactant molecules given the target product molecule. Data from USPTO-full \cite{USPTO_patent, LLaSMol} and USPTO\_500\_MT \cite{Chemformer, Mol-Instruction} are used for these tasks.

\noindent
\paragraph{Reaction Condition Prediction.} This category involves predicting the \textit{reagents}, \textit{catalysts}, and \textit{solvents} for a given reaction. We utilize extracted reaction condition information from \citet{TextReact} and re-split the reagent prediction dataset provided by \citet{Mol-Instruction} into three separate sets.

\noindent
\paragraph{Reagent Selection.} This task, also known as reagent recommendation, involves identifying the most suitable reagents for a specific chemical reaction or process. It is divided into three categories: reactant selection, ligand selection, and solvent selection. We formulate it as choosing the most suitable reagent from a list of candidates. We adopt the dataset collected from \citet{ChemLLMBench}.

\noindent
\paragraph{Reaction Type Classification.} This task aims to classify a reaction into predefined types to navigate chemical space and better understand the underlying mechanisms. We use the USPTO 1K TPL dataset from \citet{RXNFP} with 1000 labeled classes. Learned representations can also serve as reaction fingerprints, capturing fine-grained differences.

\noindent
\paragraph{Yield Regression.} This task involves estimating the amount of product (yield) obtained from a given chemical reaction. We test the model's performance on two High-Throughtput experimentation (HTE) datasets: \textit{Buchwald-Hartwig} and \textit{Suzuki-Miyaura}. Both datasets are obtained from \citet{YieldBERT}.

\noindent
% \paragraph{Remark: Prevent Data Leakage.} Early benchmark~\cite{Mol-Instruction} work failed to address data leakage issues. For instance, reagent prediction training data included retrosynthesis test results, causing bias. Additionally, random splits in SFT datasets led to significant overlap in reaction SMILES scaffolds between training and test sets, hindering real-world generalization. To prevent information leakage, we excluded reactions from the pretraining dataset that appear in the test sets of downstream tasks and eliminated data leakage between different SFT tasks. For the reaction prediction task, we reconstructed the test set using a scaffold similarity threshold to ensure a distinct distribution between training and test samples, thereby enhancing the robustness and validity of our model evaluations. For further details, please refer to the Appendix~\ref{app:data_clean}.

\paragraph{Remark: Generating an Uncontaminated and Challenging Test Set.} Data leakage is commonly observed in recent LLM studies~\cite{LanguageContamination, DataLeakage, TaskContamination}, and we have observed the same issue in early benchmarks of chemical reaction prediction~\cite{Mol-Instruction}. This issue leads to skewed evaluation and can hinder the development of truly effective models. To present a reliable chemical reaction task evaluation, we meticulously ensure no overlap between our pretraining/training datasets and testing datasets. Further, we establish a test set for the reaction prediction task by including only samples with a scaffold similarity below a certain threshold compared to the training samples. This approach separates the training and testing distributions, improving the robustness and accuracy of our evaluations. Prior benchmarks often used random splits, resulting in significant overlaps in molecular scaffolds between training and test sets, compromising the evaluation of real-world generalization. For further details, please refer to the Appendix~\ref{app:data_clean}.

% TODO: add a remark para: highlight our splitting method

% TODO: add a remark para: illustrate the eval metric

\begin{table}[t]
\centering
\small
\setlength{\tabcolsep}{1.8mm}{
\scalebox{0.78}{
\begin{tabular}{lrrrr}
\toprule
\textsc{Task} & \textsc{\# Train} & \textsc{\# Valid} & \textsc{\# Test} & \textsc{\# All} \\
% Reaction Prediction
\midrule[1.1pt]
\rowcolor[RGB]{234, 238, 234}
\multicolumn{5}{l}{\textit{Reaction Prediction}} \\
\rowcolor[RGB]{220, 230, 242}
\multicolumn{5}{l}{\textsc{Data Source:} \citet{T5Chem, LLaSMol, Mol-Instruction}} \\
Forward Prediction & 124,384 & - & 1,000 & 125,384 \\
Retrosynthesis Prediction & 124,384 & - & 1,000 & 125,384 \\
% Reaction Condition Prediction
\midrule[1.1pt]
\rowcolor[RGB]{234, 238, 234}
\multicolumn{5}{l}{\textit{Reaction Condition Prediction}} \\
\rowcolor[RGB]{220, 230, 242}
\multicolumn{5}{l}{\textsc{Data Source:} \citet{TextReact, ChemLLMBench, Mol-Instruction}} \\
Reagent Prediction & 57,162 & 6,216 & 6,378 & 69,756 \\
Catalyst Prediction & 10,232 & 1,059 & 1,015 & 12,306 \\
Solvent Prediction & 70,988 & 7,694 & 7,793 & 86,475 \\
% Reaction Condition Recommendation
\midrule[1.1pt]
\rowcolor[RGB]{234, 238, 234}
\multicolumn{5}{l}{\textit{Reaction Condition Recommendation}} \\
\rowcolor[RGB]{220, 230, 242}
\multicolumn{5}{l}{\textsc{Data Source:} \citet{ChemLLMBench}} \\
Reagent Selection & 3,955 & - & 300 & 4,255 \\
% Reaction Type Classification
\midrule[1.1pt]
\rowcolor[RGB]{234, 238, 234}
\multicolumn{5}{l}{\textit{Reaction Type Classification}} \\
\rowcolor[RGB]{220, 230, 242}
\multicolumn{5}{l}{\textsc{Data Source:} \citet{RXNFP}}\\
Reaction Type Classification & 360,379 & 40,059 & 44,511 & 445,115 \\
% Yield Prediction
\midrule[1.1pt]
\rowcolor[RGB]{234, 238, 234}
\multicolumn{5}{l}{\textit{Yield Prediction}} \\
\rowcolor[RGB]{220, 230, 242}
\multicolumn{5}{l}{\textsc{Data Source:} \citet{YieldBERT}} \\
Buchwald-Hartwig & 3,855 & - & 100 & 3,955 \\
Suzuki-Miyaura & 5,660 & - & 100 & 5,760 \\
% % Overall
% \midrule[1.1pt]
% \textbf{Overall} & 2,331,344 & 70,013 & 67,669 & 2,469,026 \\
\bottomrule
\end{tabular}
}}
\caption{\PipeName{} downstream tasks dataset statistics.}
\label{tab:dataset_downstream}

\end{table}

% \subsection{Evaluation Suite}
% % merge to 4.3
% To comprehensively assess model performance on the various reaction prediction tasks, we have developed an evaluation suite that includes metrics tailored for each task type.
% % The suite is implemented in Python and uses several popular libraries for natural language processing and cheminformatics, such as RDKit \cite{RDKit} and NLTK \cite{NLTK}. The code for the evaluation suite is provided below.
% We provide the following concrete evaluator classes:
% \begin{itemize}[leftmargin=*]
% \item \texttt{Classification}: Evaluates classification tasks using accuracy, macro-averaged F1, and micro-averaged F1 scores.
% \item \texttt{Regression}: Evaluates regression tasks using mean absolute error (MAE), mean squared error (MSE), and R$^2$ scores.
% \item \texttt{Generation}: Evaluates molecular representations (SMILES or SELFIES) using exact matching, BLEU score, Levenshtein distance, and Tanimoto similarity of molecular fingerprints (RDKit, MACCS, Morgan). It also computes the validity of the predicted molecules. Predicted SELFIES are decoded to SMILES before computing the similarity metrics.
% \end{itemize}

% By open-sourcing this evaluation suite, we hope to encourage standardized evaluation and easy reproducibility of results in the community. The modular design of the suite makes it straightforward to add support for new tasks and metrics in the future.

%% file: sec/4-exp.tex
\section{Analyzing Pre-Training Strategy and Dataset Configuration}
\label{sec:ablation_main}

In this section, we conduct experiments to evaluate the impact of different pretraining strategies and dataset configurations on downstream tasks.

\begin{figure*}[!bt]
\vskip -0.3in
    \centering
    \begin{subfigure}[b]{0.48\textwidth}
        \centering
        \includegraphics[width=\textwidth]{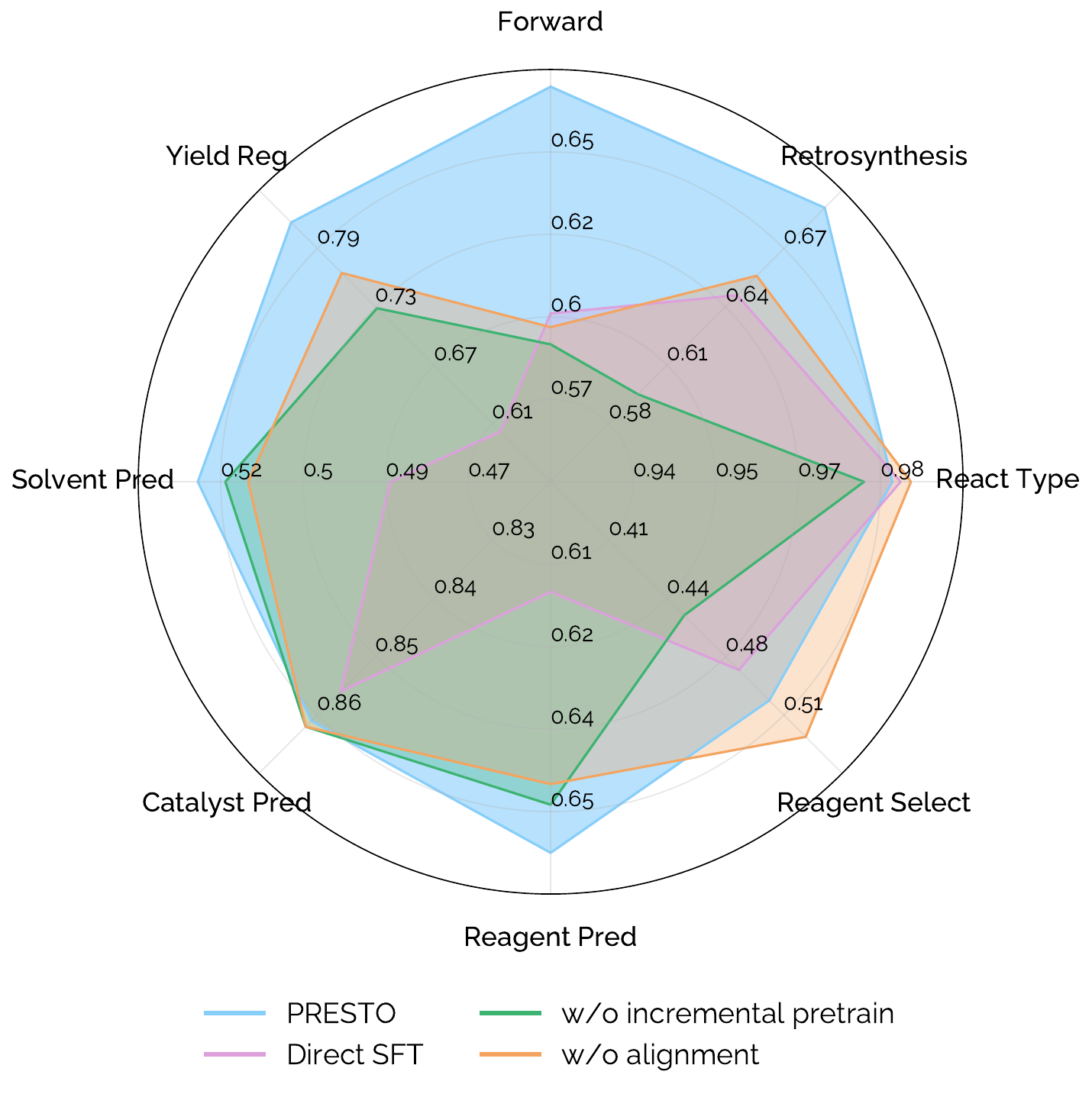}
        \caption{\textbf{Multi-stage Pretrain Strategy Ablation.}}
        % \vskip -0.1in
        \label{fig:ablation_multi_stage_perform}
    \end{subfigure}
    \hfill
    \begin{subfigure}[b]{0.48\textwidth}
        \centering
        \includegraphics[width=\textwidth]{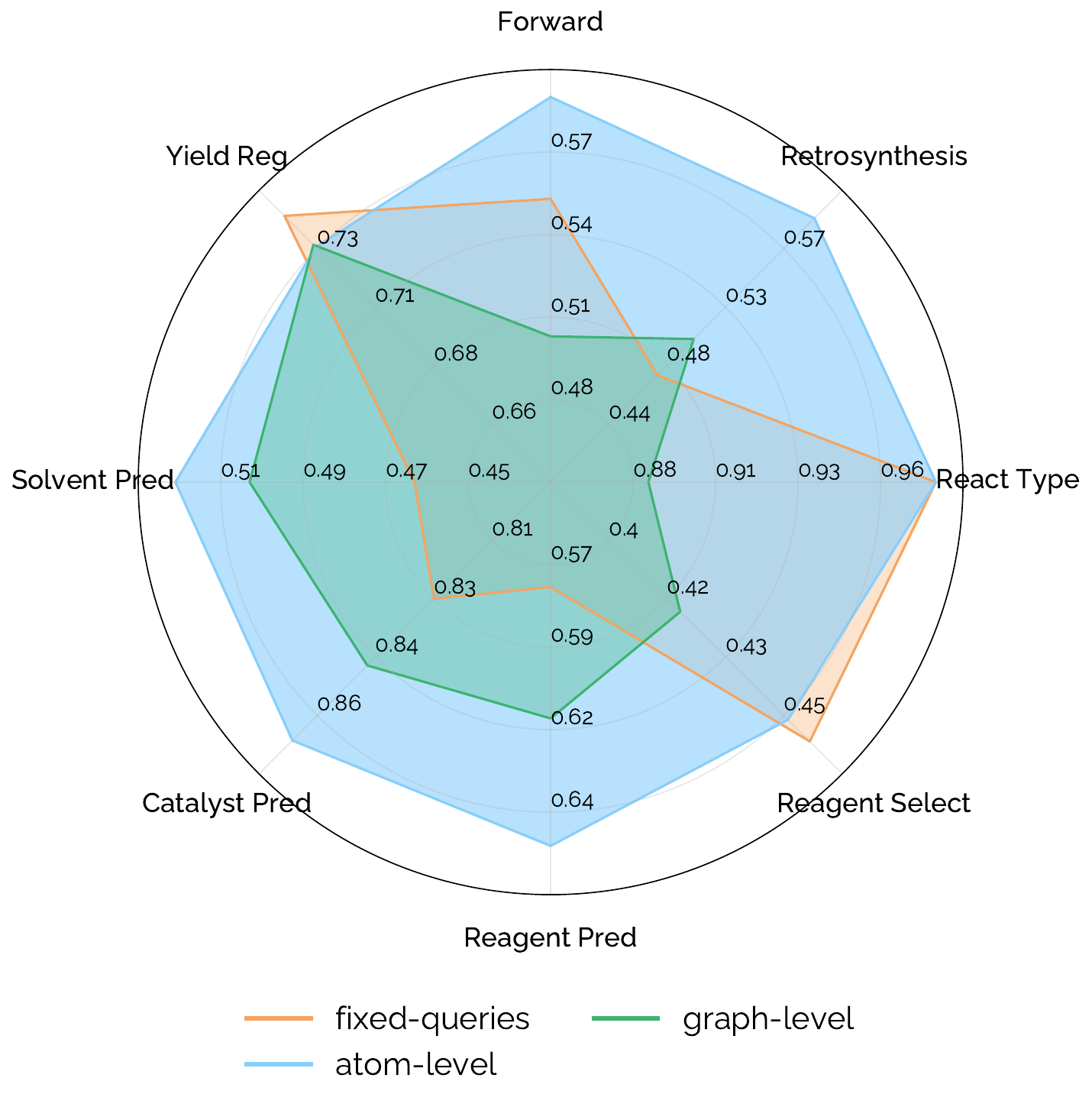}
        \caption{\textbf{Molecule Token Granularity Ablation.}}
        % \vskip -0.1in
        \label{fig:ablation_token}
    \end{subfigure}
    \vfill
    \begin{subfigure}[b]{0.48\textwidth}
        \centering
        \includegraphics[width=\textwidth]{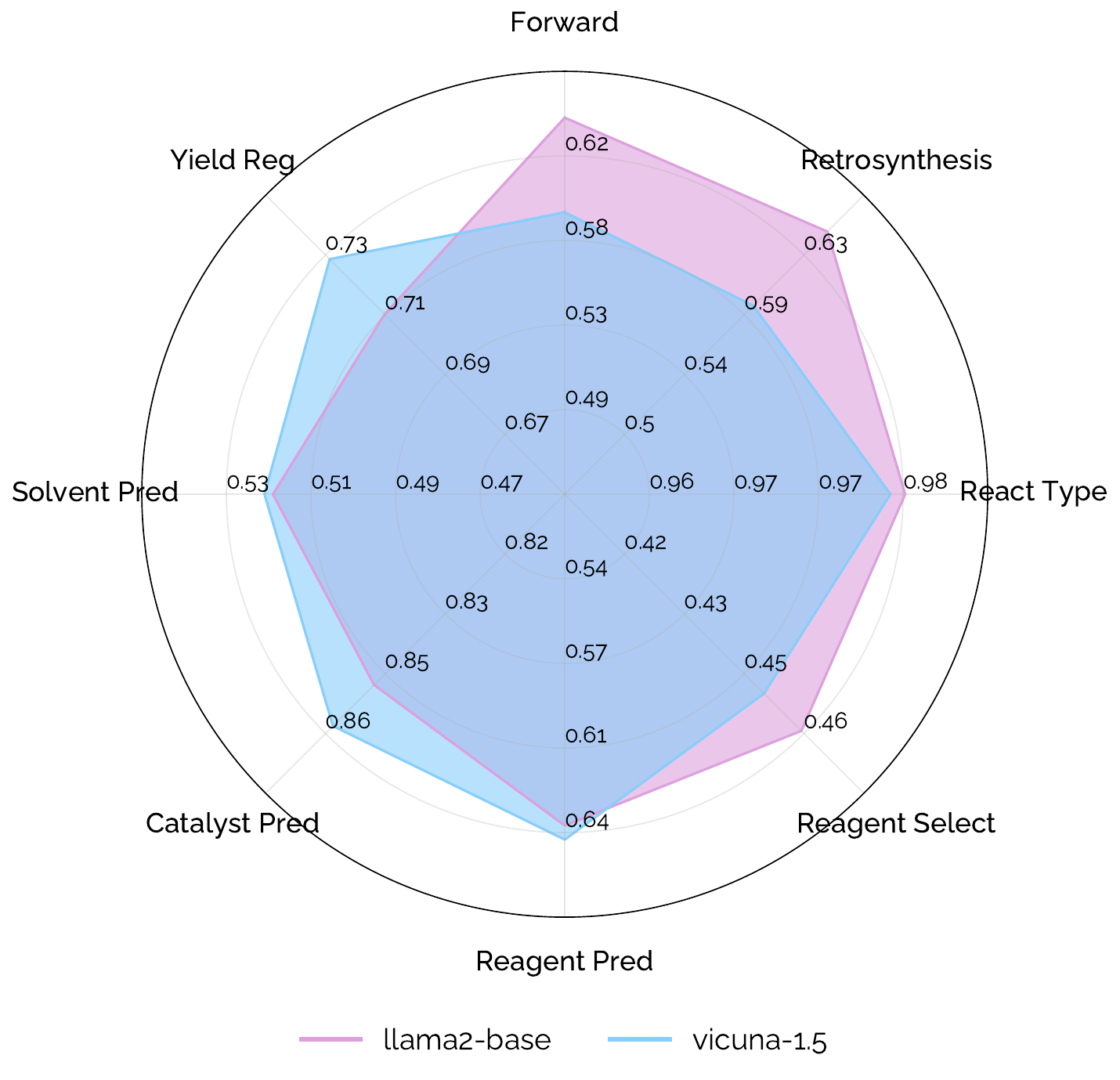}
        \caption{\textbf{Base vs. Instruct-Tuned LLMs.}}
        % \vskip -0.15in
        \label{fig:instruct_base}
    \end{subfigure}
    \hfill
    \begin{subfigure}[b]{0.48\textwidth}
        \centering
        \includegraphics[width=\textwidth]{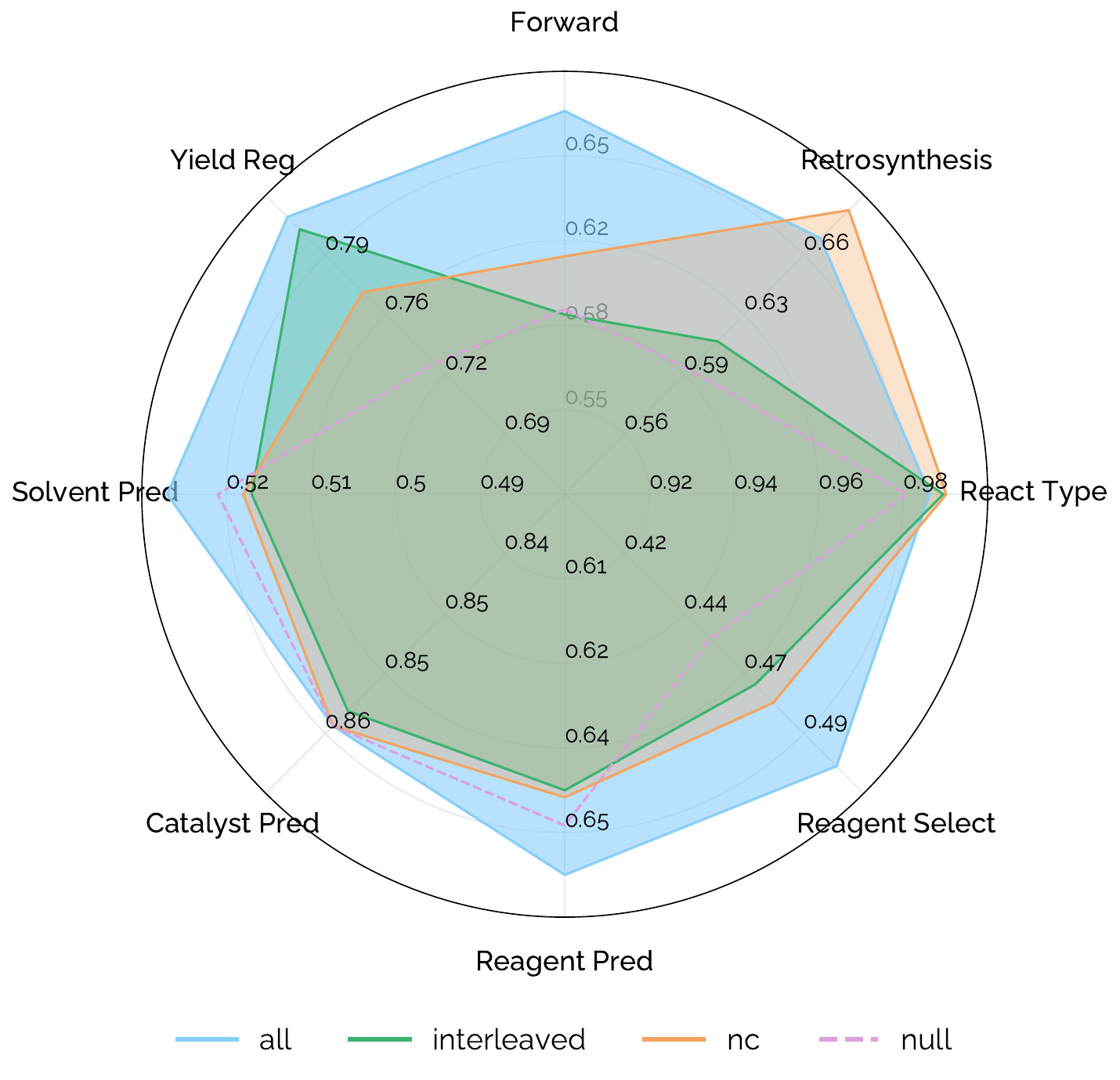}
        % \vskip -0.15in
        \caption{\textbf{PRESTO-Stage2 Dataset Configuration Ablation.}}
        \label{fig:stage-2_data}
    \end{subfigure}
    \caption{
        \textbf{Performance analysis of different pretraining strategies and dataset configurations.}
        \textbf{(a)} Ablation study on the multi-modal pretraining strategy.
        \textbf{(b)} We explore various options for the granularity of molecular encoded tokens.
        \textbf{(c)} Comparison between base (Llama-2) and instruct-tuned (Vicuna v1.5) language models. 
        \textbf{(d)} Ablation study on dataset configuration for \PipeName{} domain incremental pretraining stage.
    }
    \vskip -0.2in
\end{figure*}

\paragraph{Experimental Setting.} We use the GIN~\cite{GIN} pretrained by MoleculeSTM~\cite{MoleculeSTM} as the default graph encoder $f_M$ and a two-layer MLP as the projector $f_{\psi}$. For the base $\text{LM}_{\theta}$, we use Vicuna v1.5-7B~\cite{Vicuna} by default. We report the mean similarity measured by Morgan~\cite{Morgan}, MACCS~\cite{MACCS}, RDKit~\cite{RDKit} fingerprints for generation tasks, Top-1 accuracy for classification tasks, and $R^2$ scores for regression tasks. Detailed experimental settings are available in Appendix~\ref{app:implement_details}.

\subsection{Analyzing Pretraining Strategy}
\label{sec:ablation_pretrain}

We investigate the impact of different pretraining strategies, varying levels of molecular representation granularity, and different LLMs on the model's performance in downstream tasks.
We divide the pretraining pipeline into two stages: alignment and domain incremental pretraining, as mentioned in Section \ref{sec:3.2}. Due to the high time and computation costs of the incremental pretraining stage, we skip it unless explicitly stated otherwise.

\paragraph{Finding 1: Progressive pretraining strategy enhances downstream task performance.} As shown in Figure \ref{fig:ablation_multi_stage_perform}, \textcolor{forestpink}{Direct SFT} significantly degrades the prediction of reaction conditions and yields. This degradation occurs because the model must simultaneously learn to align different modalities and adapt to various downstream tasks, increasing the optimization difficulty. \textcolor{orange}{W/o alignment} demonstrates that the alignment stage, which acts as a warm-up for modality fusion, effectively connects molecular and language information, aiding the transition of a general-domain LLM to the chemistry domain. Additionally, \textcolor{forestgreen}{w/o incremental pretrain} underscores the importance of domain incremental pretraining in enhancing multi-graph modeling and domain knowledge adaptation.

% \begin{figure}[!hbt]
%     \centering
%     \includegraphics[width=0.5\textwidth]{figs/multi_stage_performance_ablation_radar.pdf}
%     \caption{\footnotesize{Ablation study on the multi-stage training strategy. The proposed three-stage strategy (1+2+3) achieves the best performance. Skipping stage 1 (skip-align-proj) or stage 2 (skip2) leads to a performance drop, especially on the forward synthesis and retrosynthesis tasks. Direct finetuning has the worst performance.}}
%     \label{fig:ablation_multi_stage_perform}
% \end{figure}

% \paragraph{Multi-stage strategy: ICL (PP)}
% PP(5-tasks), scaffold-8-shot(or 4? choose better one)
% Remark: Stage-2 training preserves the in-context learning ability.
% \begin{figure}[!hbt]
%     \centering
%     \includegraphics[width=0.5\textwidth]{figs/model_accuracy_8-shot_radar.pdf}
%     \caption{\footnotesize{Ablation study on the in-context learning abilities.}}
%     \label{fig:icl}
% \end{figure}

\noindent
\paragraph{Finding 2: Molecular representation granularity matters.} Drawing from prior VLMs research \citep{Prisma, VILA}, enhancing visual resolution improves downstream performance by capturing intricate details. Similarly, we utilize various granularities for molecular representation, including graph-level (a global token per graph), atom-level (each atom represented by one token), and fixed-length query-encoding \citep{3DMoLM, MolCA}. In Figure \ref{fig:ablation_token}, scaling to the atom level yields substantial improvements across all tasks compared to graph-level modeling. 
Interestingly, the query-encoding approach performs remarkably well in regression and classification tasks but severely underperforms in tasks that require generating entire molecules. We speculate that the learned queries may fail to capture fine-grained molecular structures, resulting in suboptimal performance in generating full molecules.

% \begin{figure}[!hbt]
%     \centering
%     \includegraphics[width=0.5\textwidth]{figs/token_resolution_ablation_radar.pdf}
%     \caption{\footnotesize{Ablation study on the token resolution in the adapter layers. Using the full sequence of token representations (skip2) outperforms the single vector representation (skip2-single) and the query-based representation (qformer-skip2) on most tasks, especially forward synthesis, and retrosynthesis.}}
%     \label{fig:ablation_token}
% \end{figure}

\noindent
\paragraph{Finding 3: Base and instruct-tuned LLMs demonstrate comparable capabilities.} Instruct tuning is a method to finetune base LLMs (trained for next-token prediction) to function as dialogue agents that can follow instructions more effectively. Modern VLMs research \citep{LLaVA, VILA} often use instruct-tuned models like Vicuna as the base LLMs. We evaluate the impact of instruct-tuned LLM on downstream synthetic chemistry tasks via a head-to-head comparison between a base LLM (Llama-2-7B \citep{Llama2}) and its instruct-tuned variant (Vicuna v1.5). Figure \ref{fig:instruct_base} shows that instruction-tuned LLMs slightly outperform base in reaction condition prediction and yield tasks, while base LLMs excel in forward prediction and retrosynthesis.

% \begin{figure}[!hbt]
%     \centering
%     \includegraphics[width=0.5\textwidth]{figs/base_vs_instruct_tuned_lms_radar.pdf}
%     \caption{\footnotesize{Ablation study on the base and instruct-tuned LLMs.}}
%     \label{fig:instruct_base}
% \end{figure}

\subsection{Analyzing Dataset Configuration}
\label{sec:ablation_data}

Here, we analyze the impact of dataset configurations on domain incremental pretraining.
\paragraph{Finding 4: Both interleaved data and name-conversion data play crucial roles in domain incremental pretraining.} As shown in Figure \ref{fig:stage-2_data}, relying solely on an interleaved molecule-text dataset can improve model performance in retrosynthesis, classification, and regression tasks, but the improvement is marginal. We believe this is because interleaved data lack strict molecule-text correspondence, making it difficult for the model to use the surrounding text to learn molecular syntax and semantics and recognize molecular structural patterns. Therefore, we introduce a name conversion task dataset to enhance contextual learning, which aids tasks requiring a deeper understanding of chemical entities and their functions. Experiments demonstrate that incrementally, pretraining with a blend of interleaved data and name conversion data better leverages the domain knowledge from the synthetic procedure corpus, facilitating downstream tasks.

% \begin{figure}[!hbt]
%     \centering
%     \includegraphics[width=0.5\textwidth]{figs/stage2_pretrain_dataset_composition_radar.pdf}
%     \caption{\footnotesize{Ablation study on the dataset composition for stage 2 pretraining. A combination of all data improves the ability for forward synthesis and retrosynthesis.}}
%     % all abbr explain
%     \label{fig:stage-2_data}
% \end{figure}

\input{sec/tab-performance}

\section{Comparison with the State-of-the-arts}
We integrate the above findings to inform our \ModelName{} framework at the 7B parameter scale. We present results comparing \ModelName{} with previous domain expert models~\citep{Chemformer, MolecularTransformer, Retroformer, RXNFP, ContraGIN, DRFP, DFT, UAGNN, YieldBERT} and other LLM-based methods~\citep{Mol-Instruction, nach0, Text+ChemT5, LLaSMol, Galactica, ChemDFM, T5Chem}.

Table \ref{tab:chemical_reaction} presents the performances for generation tasks. We report commonly used metrics in the MTM domain, including Exact Match, BLEU~\cite{BLEU}, Levenshtein distance, Validity, and fingerprint similarities (RDKit, MACCS, and Morgan). Table \ref{tab:other} reports on regression and classification tasks, evaluating metrics such as Accuracy, Confusion Entropy of the confusion matrix (CEN), Matthews Correlation Coefficient (MCC), and $R^2$ scores. Results show that \ModelName{} outperforms the baseline LLMs across all downstream tasks and narrows the gap with domain expert models. These improvements highlight the effectiveness of our proposed progressive pretraining strategy and comprehensive analytical design. However, it is noteworthy that there is still room for improvement in validity. Future efforts could involve replacing SMILES with SELFIES~\cite{SELFIES} to enhance robustness in representation.
% NOTE: TextReact主要是retrieve model, 无法参与评价condition prediction

%% file: sec/tab-performance.tex
\begin{table*}[t]
\vskip -0.4in
\centering
\small
\setlength{\tabcolsep}{2.2mm}{
\scalebox{0.77}{
\begin{tabular}{lccccccc}
\toprule
\textsc{Model}
&\textsc{Exact}$\uparrow$  & \textsc{BLEU}$\uparrow$  & \textsc{Levenshtein}$\downarrow$  & \textsc{RDK FTS}$\uparrow$  & \textsc{MACCS FTS}$\uparrow$ & \textsc{Morgan FTS}$\uparrow$ & \textsc{Validity}$\uparrow$ \\

% Forward Prediction
\midrule[1.1pt]
\rowcolor[RGB]{234, 238, 234}
\multicolumn{8}{l}{\textit{Forward Reaction Prediction}} \\
% \color{gray} Chemformer$^*$~\cite{Chemformer} & \color{gray}0.502	&\color{gray}0.834	&\color{gray}6.878	&\color{gray}0.756	&\color{gray}0.821	&\color{gray}0.739	&\color{gray}0.986 \\
\color{gray} Chemformer$^*$~\cite{Chemformer} & \color{gray}0.372 & \color{gray}0.824 & \color{gray}8.097 & \color{gray}0.755 & \color{gray}0.820 & \color{gray}0.717 & \color{gray}0.994 \\
\color{gray} MoleculeTransformers$^*$~\cite{MolecularTransformer} & \color{gray}0.313 & \color{gray}0.663 & \color{gray}11.735 & \color{gray}0.549 & \color{gray}0.619 & \color{gray}0.532 & \color{gray}0.925 \\
Mol-Instruction~\cite{Mol-Instruction} & 0.065 & 0.428 & 24.076 & 0.260 & 0.430 & 0.249 & \textbf{0.999} \\
LLama2-7b$^*$~\cite{Llama2} &0.251	&0.658	&13.167	&0.533	&0.630	&0.512	&0.940 \\
Vicuna v1.5-7b$^*$~\cite{Vicuna} &0.250	&0.659	&12.506	&0.513	&0.600	&0.495	&0.903 \\
LlaSMol-Mistral~\cite{LLaSMol} & 0.055	&0.750	&15.558	&0.221	&0.471	&0.202	&0.788 \\
nach0-base~\cite{nach0} &0.331	&\textbf{0.857}	&13.108	&0.628	&0.709	&0.594	&0.977 \\
Text+Chem T5~\cite{Text+ChemT5} &0.236	&0.750	&13.631	&0.523	&0.630	&0.505	&0.967 \\
T5Chem~\cite{T5Chem} &0.313	&0.703	&13.632	&0.535	&0.616	&0.520	&0.965 \\
\midrule
\textbf{\ModelName} & \textbf{0.355} & 0.836 & \textbf{10.647} &\textbf{0.646}	&\textbf{0.726}	&\textbf{0.624}	&0.973 \\

% Retrosynthesis Prediction
\midrule[1.1pt]
\rowcolor[RGB]{234, 238, 234}
\multicolumn{8}{l}{\textit{Retrosynthesis Prediction}} \\
% \color{gray} Chemformer$^*$ & \color{gray}0.360	&\color{gray}0.791	&\color{gray}14.114	&\color{gray}0.738	&\color{gray}0.826	&\color{gray}0.698	&\color{gray}0.995 \\
\color{gray} Chemformer$^*$ & \color{gray}0.011 & \color{gray}0.611 & \color{gray}21.073 & \color{gray}0.659 & \color{gray}0.730 & \color{gray}0.574 & \color{gray}0.998 \\
\color{gray} Retroformer$^*$ \citep{Retroformer} & \color{gray}0.273 & \color{gray}0.769 & \color{gray}14.768 & \color{gray}0.690 & \color{gray}0.782 & \color{gray}0.647 & \color{gray}0.952 \\
Mol-Instruction & 0.039 & 0.395 & 31.611 & 0.279 & 0.478 & 0.26 & \textbf{1.0} \\
LLama2-7b$^*$ &0.220	&0.754	&15.695	&0.649	&0.747	&0.608	&0.933 \\
Vicuna v1.5-7b$^*$&0.220	&0.756	&15.692	&0.658	&0.758	&0.616	&0.943 \\
LlaSMol-Mistral & 0.010	&0.694	&19.719	&0.148	&0.317	&0.119	&0.530 \\
nach0-base &0.173	&0.854	&18.883	&0.574	&0.668	&0.515	&0.892 \\
Text+Chem T5 &0.042 &0.620 &13.952 &0.261 &0.281 &0.206 &0.345 \\
T5Chem &0.208	&0.725	&17.278	&0.595	&0.662	&0.566	&0.994\\
\midrule
\textbf{\ModelName} & \textbf{0.275} & \textbf{0.902} & \textbf{14.433} & \textbf{0.655}	& \textbf{0.747} & \textbf{0.619}	& 0.980 \\

% Reaction Condition Prediction (Reagent)
\midrule[1.1pt]
\rowcolor[RGB]{234, 238, 234}
\multicolumn{8}{l}{\textit{Reaction Condition Prediction (Reagent)}} \\
LLama2-7b$^*$ &0.312	&0.564	&9.058	&0.560	&0.575	&0.466	&\textbf{1.0} \\
Vicuna v1.5-7b$^*$ &0.315	&0.585	&8.664	&0.576	&0.587	&0.473	&\textbf{1.0} \\
nach0-base &0.001	&0.172	&34.212	&0.053	&0.134	&0.039	&0.932\\
Mol-Instruction &0.0	&0.219	&27.108	&0.034	&0.098	&0.030	&\textbf{1.0}\\
T5Chem &0.019	&0.559	&11.044	&0.366	&0.461	&0.374	&0.994\\
\midrule
\textbf{\ModelName} & \textbf{0.458}	&\textbf{0.776}	&\textbf{6.206}	&\textbf{0.678}	&\textbf{0.683}	&\textbf{0.601}	&0.999\\

% Reaction Condition Prediction (Catalyst)
\midrule[1.1pt]
\rowcolor[RGB]{234, 238, 234}
\multicolumn{8}{l}{\textit{Reaction Condition Prediction (Catalyst)}} \\
LLama2-7b$^*$ &0.680	&0.720	&2.545	&0.882	&0.868	&0.687	&\textbf{1.0} \\
Vicuna v1.5-7b$^*$ &0.685	&0.703	&2.451	&0.883	&0.869	&0.692	&\textbf{1.0} \\
nach0-base &0.0	&0.072	&36.442	&0.129	&0.055	&0.009	&0.849 \\
Mol-Instruction &0.0	&0.110	&28.424	&0.031	&0.045	&0.015	&0.999\\
T5Chem &0.022	&0.346	&13.408	&0.146	&0.268	&0.200	&0.996\\
\midrule
\textbf{\ModelName}       & \textbf{0.768}	&\textbf{0.814}	&\textbf{1.755}	&\textbf{0.914}	&\textbf{0.895}	&\textbf{0.774}	&\textbf{1.0}\\

% Reaction Condition Prediction (Solvent)
\midrule[1.1pt]
\rowcolor[RGB]{234, 238, 234}
\multicolumn{8}{l}{\textit{Reaction Condition Prediction (Solvent)}} \\
LLama2-7b$^*$ &0.311	&0.462	&3.819	&0.452	&0.48	&0.417	&\textbf{1.0} \\
Vicuna v1.5-7b$^*$ &0.320	&0.436	&3.809	&0.459	&0.486	&0.427	&\textbf{1.0} \\
nach0-base &0.0	&0.072	&36.442	&0.129	&0.055	&0.009	&0.849\\
Mol-Instruction &0.0	&0.155	&25.117	&0.030	&0.122	&0.035	&\textbf{1.0}\\
T5Chem &0.083	&0.311	&16.224	&0.458	&0.424	&0.397	&0.995\\
\midrule
\textbf{\ModelName}  & \textbf{0.419} & \textbf{0.695}	& \textbf{2.758} & \textbf{0.529}	& \textbf{0.547} & \textbf{0.506}	& 0.912\\

\bottomrule
\end{tabular}
}}
\vskip -0.1in
\caption{\small{
Comparison of various models on forward reaction prediction, retrosynthesis prediction, and reaction condition prediction tasks.  \textcolor{gray}{Model} indicates a domain expert method, and $^*$ denotes our re-implementation. }
}
\label{tab:chemical_reaction}
\end{table*}
% NOTE: we exclude 

\begin{table*}[!htbp]
\centering
\begin{tabular}{@{}c@{\hspace{0.5cm}}c@{}c@{}}
  % first subtable
  \begin{subtable}[t]{0.35\linewidth} % Adjusted width
    \setlength{\tabcolsep}{2.8 pt}
    \renewcommand\arraystretch{1.0}
    \centering
    \scalebox{0.55}{
    \begin{tabular}{lccc}
        \toprule
        \textsc{Method} &\textsc{Reactant} &\textsc{Solvent} &\textsc{Ligand}\\
        \midrule[1.1pt]
        \rowcolor[RGB]{234, 238, 234} \multicolumn{4}{l}{\textit{Reagent Selection}} \\
        \rowcolor[RGB]{255, 255, 255}
        LLama2-7b$^*$& 0.670 & 0.550 & 0.010\\
        Vicuna v1.5-7b$^*$& 0.690 & 0.580 & 0.440\\
        GPT-4$^\dagger$& 0.299 & 0.526 & \textbf{0.534} \\
        GAL-30B$^\dagger$~\cite{Galactica}& 0.107 & 0.104 & 0.030 \\
        LLama2-13b-chat$^\dagger$& 0.145 & 0.050 & 0.284 \\
        ChemDFM-13b~\cite{ChemDFM} & 0.240 & 0.120 & 0.350 \\
        \midrule
        \textbf{\ModelName{}}& \textbf{0.780} & \textbf{0.630} & 0.520 \\
        \bottomrule
    \end{tabular}
    }
  \end{subtable}
  &
  % second subtable
  \begin{subtable}[t]{0.35\linewidth} % Adjusted width
    \setlength{\tabcolsep}{2.8 pt}
    \renewcommand\arraystretch{1.0}
    \centering
    \scalebox{0.55}{
    \begin{tabular}{lccc}
        \toprule
        \textsc{Method} &\textsc{Acc}$\uparrow$ & \textsc{CEN}$\downarrow$ & \textsc{MCC}$\uparrow$ \\
        \midrule[1.1pt]
        \rowcolor[RGB]{234, 238, 234} \multicolumn{4}{l}{\textit{Reaction Type Classification}} \\
        \rowcolor[RGB]{255, 255, 255}
        \color{gray}BERT classifier~\cite{RXNFP} & \color{gray} 0.989 & \color{gray}0.006 & \color{gray}0.989\\
        \color{gray}ContraGIN~\cite{ContraGIN} & \color{gray} 0.993 & \color{gray}0.001 & \color{gray}0.993 \\
        \color{gray}DRFP~\cite{DRFP} & \color{gray} 0.977 & \color{gray}0.011 & \color{gray}0.977\\
        T5Chem & \textbf{0.995} & \textbf{0.003} & \textbf{0.995}\\
        LLama2-7b$^*$ & 0.804 & 0.079  & 0.803 \\
        Vicuna v1.5-7b$^*$ & 0.888 & 0.048 & 0.887\\
        \midrule
        \textbf{\ModelName{}} & 0.991 & 0.004 &0.991\\
        \bottomrule
    \end{tabular}
    }
  \end{subtable}
  &
  % third subtable
  \begin{subtable}[t]{0.30\linewidth} % Adjusted width
    \setlength{\tabcolsep}{2.8 pt}
    \renewcommand\arraystretch{1.0}
    \centering
    \scalebox{0.55}{
    \begin{tabular}{lcc}
        \toprule
        \textsc{Method} &\textsc{B-H} &\textsc{S-M} \\
        \midrule[1.1pt]
        \rowcolor[RGB]{234, 238, 234} \multicolumn{3}{l}{\textit{Yield Regression}} \\
        \rowcolor[RGB]{255, 255, 255}
        \color{gray}DFT~\cite{DFT} & \color{gray}0.920 & \color{gray}- \\
        \color{gray}UAGNN~\cite{UAGNN} & \color{gray}0.969 & \color{gray}0.884 \\
        \color{gray}YieldBERT~\cite{YieldBERT} & \color{gray}0.950 & \color{gray}0.815 \\
        T5Chem & \textbf{0.970} & - \\
        LLama2-7b$^*$ & -0.476 & 0.121 \\
        Vicuna v1.5-7b$^*$ & -0.131 & 0.151 \\
        \midrule
        \textbf{\ModelName{}} & 0.944 & \textbf{0.652} \\
        \bottomrule
    \end{tabular}
    }
  \end{subtable}
\end{tabular}
\vskip -0.1in
\caption{\small{
Comparison with baselines on reagent selection, reaction type classification, and yield regression tasks. $^\dagger$ denotes results from~\cite{ChemDFM}. For reagent selection, we report the result in top-1 accuracy except for \textsc{Ligand Selection}, where we report the top 50\% accuracy. For yield regression, we report the $R^2$ score.
}}
\label{tab:other}
\vskip -0.2in
\end{table*}

%% file: sec/5-conclusion.tex
\section{Conclusion and Future Work}
\vskip -0.1in
This study explores integrating multimodal LLMs into synthetic chemistry tasks to overcome the molecule-text modality gap. We highlight the importance of multi-graph datasets and progressive pretraining methods, showing significant improvements in reaction predictions and synthetic chemistry tasks. As a result, we introduce \PipeName{}, which outperforms baseline LLMs.

Meanwhile, current multimodal molecule models are limited to generating only 1D sequences. As a potential direction, we envision developing models capable of producing comprehensive molecular representations (i.e., 2D, 3D). Future research could also expand the diversity of datasets to include more molecular structures and improve the LLM's capability for dialogue. We aim to advance the fields of synthetic chemistry and compound discovery, ultimately creating a more powerful and versatile assistant for chemists.

% \clearpage
\section*{Limitations}
\label{sec:Limitations} 
Despite the significant advancements achieved by \PipeName{}, several limitations remain. Firstly, we did not conduct ablation studies on additional molecular modalities, such as 3D structure information, nor did we explore whether combining different modalities could further enhance molecular representations and improve downstream performance. Secondly, we observed that the model's ability to answer general domain questions declined as domain-specific finetuning (SFT) progressed. Future training should consider integrating general domain SFT datasets to prevent the forgetting issue. Lastly, our base LLM is a general-domain model, and the fields of chemistry and molecular science lack specialized LLMs with parameter scales comparable to models like LLaMA. This limitation restricts the coverage and application of domain-specific knowledge, underscoring the need to develop larger, more versatile domain-specific LLMs for enhanced performance.

%%% %%%
\section*{Potential Risks}
\label{sec:Ethical}
The use of AI in synthetic chemistry carries several potential risks. One major concern is the possibility of misuse to produce dangerous or illicit substances, posing significant safety and ethical challenges. Additionally, inaccuracies in the generated content could lead to hazardous chemical reactions if not carefully verified, potentially causing harm or equipment damage. Over-reliance on AI-generated synthesis procedures without proper validation increases the risk of accidents and unsafe practices. Strict oversight and robust ethical guidelines are essential to mitigate these risks and ensure safe application.

%% file: sec/suppl.tex
\clearpage

\input{sec/appendix/a-data}

\input{sec/appendix/b-details}

\input{sec/appendix/c-ablation-SFT}

\input{sec/appendix/d-template}

\input{sec/appendix/e-examples}

%% file: sec/appendix/a-data.tex
\section{Data Collection}
\label{app:data_collection}

All the SMILES strings are canonicalized using RDKit \cite{RDKit} to ensure a standard representation. We apply additional data cleaning steps, such as removing invalid SMILES and handling duplicate entries.

\subsection{Data Cleaning}
\label{app:data_clean}
\noindent
\paragraph{Data leakage in prior works.} 
Our experiments identified data leakage issues in the previous popular benchmark study Mol-Instruction~\cite{Mol-Instruction}. For example, in the retrosynthesis prediction task, we compared reactions in the train and test splits after canonicalizing SMILES and found that 72 chemical reactions in the test split also appeared in the train split. Moreover, in the reagent prediction task, 884 reactions in the train split were identical to those in the test split of the retrosynthesis prediction task. Additionally, the study employed a random split method for train and test sets, which resulted in significant molecular scaffold similarities (Fingerprint Tanimoto Similarity avg $\sim$ 0.8) between the reactions in the train and test splits. Consequently, the test results on this benchmark lack generalizability for real-world applications.

\begin{figure*}[!hbt]
    \centering
    \includegraphics[width=\textwidth]{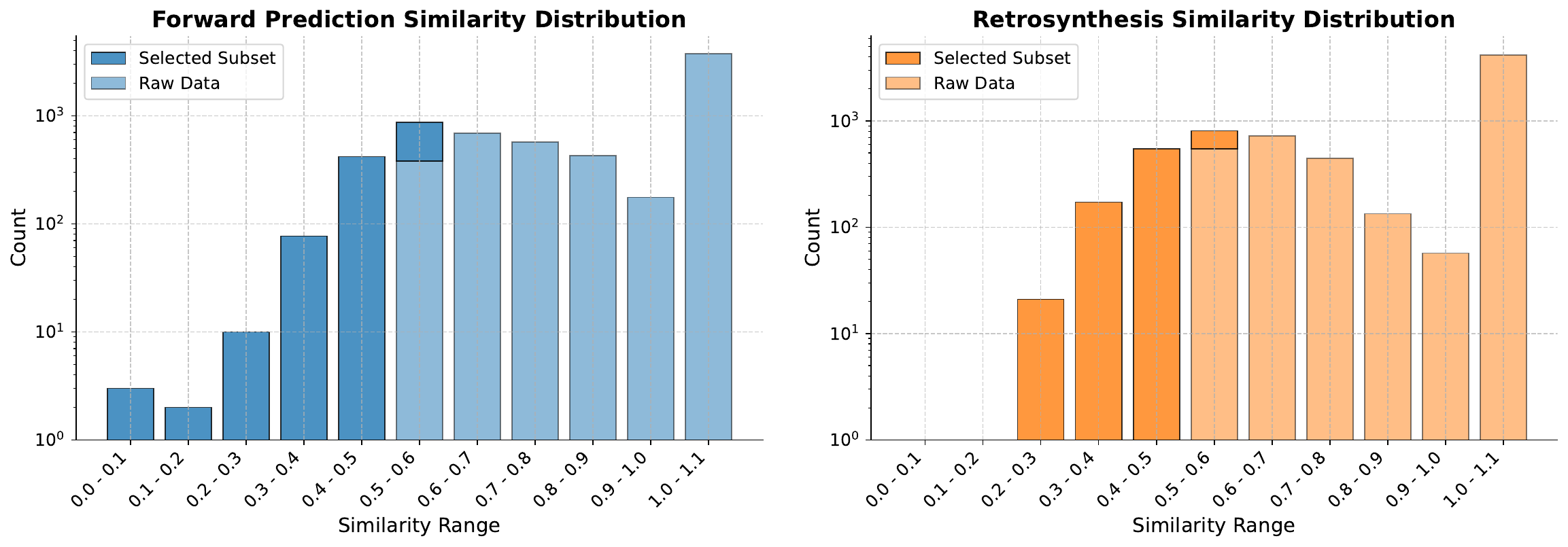}
    \caption{\textbf{Comparison of similarity distributions for reaction prediction datasets.} The plots show the count of scaffolds within each similarity range for the full test datasets provided in \citet{LLaSMol} and \citet{Mol-Instruction} (raw data, lighter shade) and the selected subsets of 1000 scaffolds with the lowest similarities (darker shade).}
    \label{fig:scaffold_comparison}
\end{figure*}

\noindent
\paragraph{Our non-overlapping, scaffold-based dataset splits.} 
When splitting the dataset, we followed two principles: (1) Ensure that chemical reactions in the test splits of all downstream synthetic chemistry tasks do not appear in any train datasets, including both the pretraining and SFT train datasets; (2) Resample the test set based on a scaffold splitting approach, using a scaffold similarity threshold (Fingerprint Tanimoto Similarity set between 0.5 and 0.6). The number of samples was maintained consistent with the Mol-Instruction test set, with additional samples selected from the LlaSMol~\citep{LLaSMol} test set. 
Figure~\ref{fig:scaffold_comparison} illustrates the scaffold similarity distribution of reaction SMILES between previous works and our resampled test set.

\subsection{Data Collection and Preprocessing of \PipeName{}}
\label{app:data_interleaved}
In this section, we provide details on the data collection and preprocessing procedures for PRESTO two pretraining stages.
\noindent
\paragraph{PubChem Caption Dataset for Mol-Text Alignment.} We constructed a molecule caption dataset to enable the LLM to integrate molecule structure information and biomolecular domain knowledge during the initial alignment phase. Using the PubChem~\cite{PubChem} database as the data source, we followed the construction procedures outlined in \citet{MoleculeSTM}. For each molecule, we used the ``description'' field from its annotation page as the corresponding text description. This resulted in a dataset of 326,675 molecule-text pairs.

\begin{figure*}[!hbt]
    \centering
    \begin{subfigure}[b]{0.47\textwidth}
        \centering
        \includegraphics[width=\textwidth]{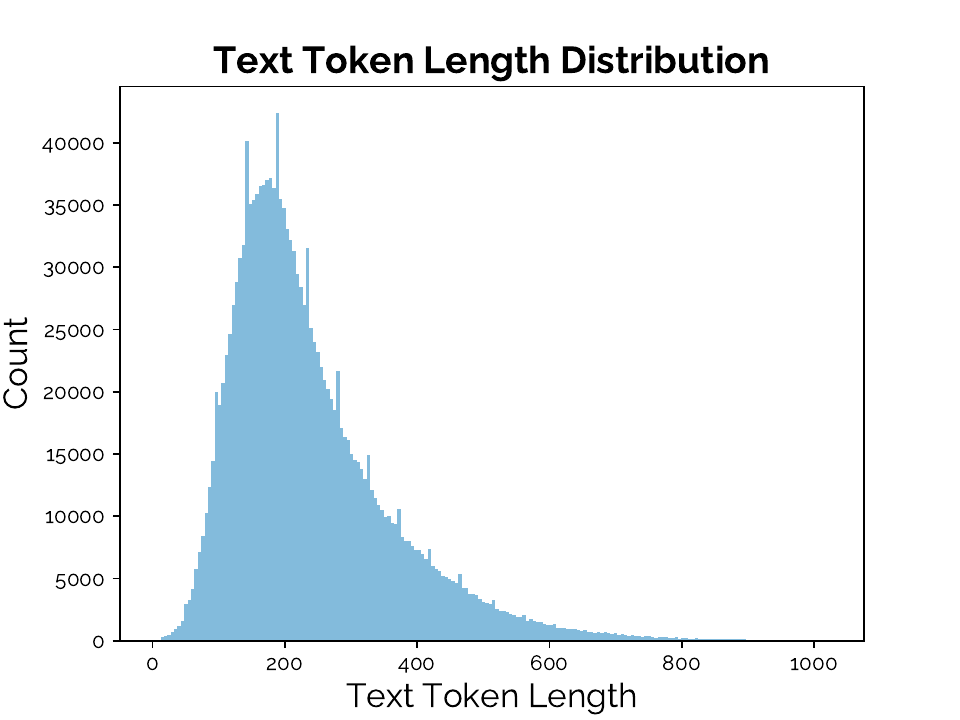}
        % \caption{}
        % \vskip -0.1in
        \label{fig:USPTO_text_token_len}
    \end{subfigure}
    \begin{subfigure}[b]{0.47\textwidth}
        \centering
        \includegraphics[width=\textwidth]{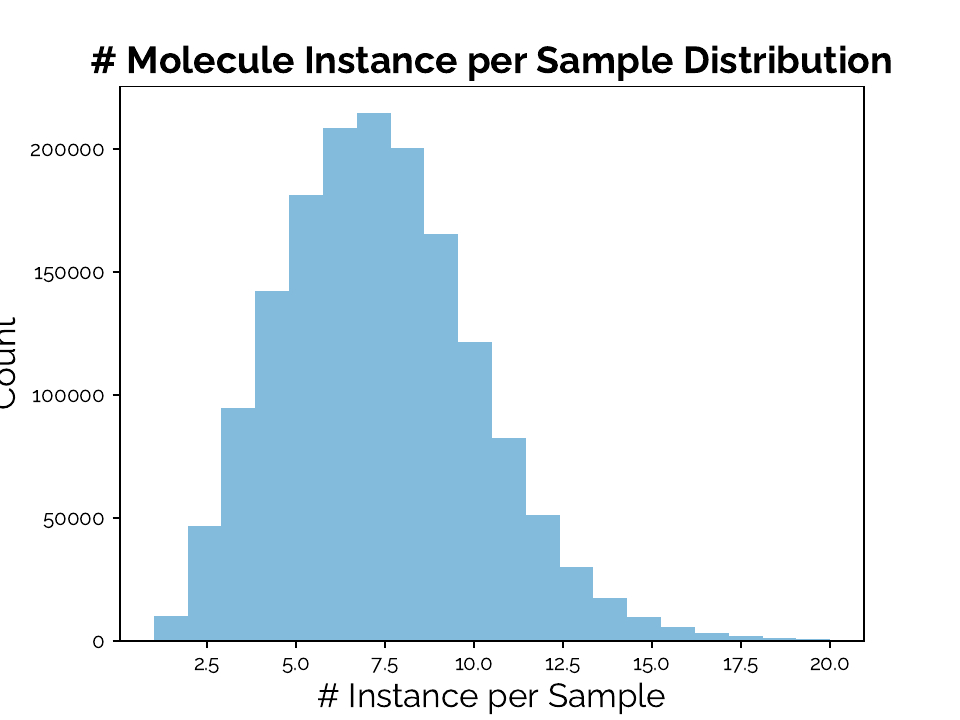}
        % \caption{\textbf{Molecule Token Granularity Ablation.}}
        % 
\label{fig:USPTO_mol_instance_num}
    \end{subfigure}
    \vskip -0.2in
    \caption{\textbf{Statistics of the Interleaved Molecule-Text Dataset.}}
    \label{fig:interleaved_stats}
\end{figure*}

\noindent
\paragraph{Interleaved Dataset for Domain Incremental Pretrain.} We compiled the interleaved molecule-text dataset primarily from USPTO-Applications~\cite{USPTO_patent}, consisting of approximately 2 million reactions and their corresponding application records published by USPTO between 2001 and September 2016. Raw XML files were downloaded, and key information for each reaction, including chemical reaction equations and textual descriptions of experimental procedures, was extracted. Following initial deduplication and filtering procedures outlined in \cite{Wang2023GenericIR}, we initially collected 1,593,329 procedure samples. Subsequently, we proceeded with two main preprocessing steps:

\begin{itemize}[leftmargin=10pt]
    \item Entity Recognition: We used the Named Entity Recognition tool BERN2 \cite{BERN2} to extract molecule entities from procedure paragraphs, retaining samples containing identifiable molecule entities. All extracted molecules' IUPAC names were then converted to SMILES format, suitable for further encoding into 2D molecular graphs. After this step, 1,592,462 samples remained.
    \item Removal of samples with excessive molecule entities and sequence length: To manage token space and prevent overly long sequences, samples containing more than 20 entities (filtering out 1,556 samples) and text sequences exceeding 1024 tokens (filtering out 2,197 samples) were removed. Finally, our constructed interleaved dataset comprises 1,588,709 samples, encompassing over 342,401 unique molecules. The statistics of the interleaved molecule-text dataset are shown in Figure~\ref{fig:interleaved_stats}.
\end{itemize}

\noindent
\paragraph{Name Conversion Dataset for Domain Incremental Pretrain.} 
We collected molecule entries from PubChem \citep{PubChem} and utilized the existing dataset from LLaSMol \citep{LLaSMol}. LLaSMol originally presents four tasks: SMILES to Formula, SMILES to IUPAC name, IUPAC name to SMILES, and IUPAC name to Formula. We retained the latter two tasks as text-only data. To integrate molecule graph tokens into \PipeName{}, we replaced SMILES with graph representations using \citet{RDKit}, creating two new tasks: Molecule Graph to Formula and Molecule Graph to IUPAC. Additionally, we derived a fifth task, Molecule Graph to SMILES, directly from the \citet{PubChem} molecule entries by parsing the SMILES into graph representations similarly.

\begin{table*}[!hbt]
\scriptsize
\tabcolsep=0.1cm
\centering
\resizebox{0.95\textwidth}{!}{%
\begin{tabular}{lccccccccc}
\toprule
\multirow{2}{*}{\textbf{Method}} & \multirow{2}{*}{\textbf{Forward}} & \multirow{2}{*}{\textbf{Retro}} & \multicolumn{4}{c}{\textbf{Reaction Condition Pred}} & \multirow{2}{*}{\begin{tabular}[c]{@{}c@{}}\textbf{Reagent}\\ 
\textbf{Recommend}\end{tabular}} & \multirow{2}{*}{\begin{tabular}[c]{@{}c@{}}\textbf{Reaction}\\ \textbf{Type}\end{tabular}} & \multirow{2}{*}{\textbf{Yield}} \\ \cmidrule(lr){4-7} 
 &  &  & All & Reagent & Catalyst & Solvent &  &  &  \\ \midrule
T5Chem \cite{T5Chem} & \textcolor{teal}{$\checkmark$} & \textcolor{teal}{$\checkmark$} & \textcolor{teal}{$\checkmark$} & \textcolor{red}{\ding{55}} & \textcolor{red}{\ding{55}} & \textcolor{red}{\ding{55}} & \textcolor{red}{\ding{55}} & \textcolor{teal}{$\checkmark$} & \textcolor{teal}{$\checkmark$} \\ 
Text+ChemT5 \cite{Text+ChemT5} & \textcolor{teal}{$\checkmark$} & \textcolor{teal}{$\checkmark$} & \textcolor{red}{\ding{55}} & \textcolor{red}{\ding{55}} & \textcolor{red}{\ding{55}} & \textcolor{red}{\ding{55}} &\textcolor{red}{\ding{55}} & \textcolor{red}{\ding{55}} & \textcolor{red}{\ding{55}} \\ 
TextReact \cite{TextReact} & \textcolor{red}{\ding{55}} & \textcolor{teal}{$\checkmark$} & \textcolor{teal}{$\checkmark$} & \textcolor{teal}{$\checkmark$} & \textcolor{teal}{$\checkmark$} & \textcolor{teal}{$\checkmark$} & \textcolor{teal}{$\checkmark$} & \textcolor{red}{\ding{55}} & \textcolor{red}{\ding{55}} \\
ChemDFM \cite{ChemDFM} & \textcolor{teal}{$\checkmark$} & \textcolor{teal}{$\checkmark$} & \textcolor{teal}{$\checkmark$} & \textcolor{red}{\ding{55}} & \textcolor{red}{\ding{55}} & \textcolor{teal}{$\checkmark$} & \textcolor{teal}{$\checkmark$} & \textcolor{red}{\ding{55}} & \textcolor{teal}{$\checkmark$} \\
Mol-Instruction \cite{Mol-Instruction} & \textcolor{teal}{$\checkmark$} & \textcolor{teal}{$\checkmark$} & \textcolor{teal}{$\checkmark$} & \textcolor{red}{\ding{55}} & \textcolor{red}{\ding{55}} & \textcolor{red}{\ding{55}} & \textcolor{red}{\ding{55}} & \textcolor{red}{\ding{55}} & \textcolor{red}{\ding{55}} \\
LlaSMol \cite{LLaSMol} & \textcolor{teal}{$\checkmark$} & \textcolor{teal}{$\checkmark$} & \textcolor{red}{\ding{55}} & \textcolor{red}{\ding{55}} & \textcolor{red}{\ding{55}} & \textcolor{red}{\ding{55}} & \textcolor{red}{\ding{55}} & \textcolor{red}{\ding{55}} & \textcolor{red}{\ding{55}} \\
BioT5+ \cite{BioT5+} & \textcolor{teal}{$\checkmark$} & \textcolor{teal}{$\checkmark$} & \textcolor{teal}{$\checkmark$} & \textcolor{red}{\ding{55}} & \textcolor{red}{\ding{55}} & \textcolor{red}{\ding{55}} & \textcolor{red}{\ding{55}} & \textcolor{red}{\ding{55}} & \textcolor{red}{\ding{55}} \\
InstructMol \cite{InstructMol} & \textcolor{teal}{$\checkmark$} & \textcolor{teal}{$\checkmark$} & \textcolor{teal}{$\checkmark$} & \textcolor{red}{\ding{55}} & \textcolor{red}{\ding{55}} & \textcolor{red}{\ding{55}} & \textcolor{red}{\ding{55}} & \textcolor{red}{\ding{55}} & \textcolor{red}{\ding{55}} \\
nach0 \cite{nach0} & \textcolor{teal}{$\checkmark$} & \textcolor{teal}{$\checkmark$} & \textcolor{teal}{$\checkmark$} & \textcolor{red}{\ding{55}} & \textcolor{red}{\ding{55}} & \textcolor{red}{\ding{55}} & \textcolor{red}{\ding{55}} & \textcolor{red}{\ding{55}} & \textcolor{red}{\ding{55}} \\
\ModelName  & \textcolor{teal}{$\checkmark$} & \textcolor{teal}{$\checkmark$} & \textcolor{teal}{$\checkmark$}  & \textcolor{teal}{$\checkmark$} & \textcolor{teal}{$\checkmark$} & \textcolor{teal}{$\checkmark$} & \textcolor{teal}{$\checkmark$} & \textcolor{teal}{$\checkmark$} & \textcolor{teal}{$\checkmark$}\\
\bottomrule
\end{tabular}%
}
\caption{
\textbf{Comparison of various models across different chemical reaction prediction tasks.} The table summarizes the capabilities of each method in forward reaction prediction, retrosynthesis prediction, reaction condition prediction (overall, reagent, catalyst, and solvent), reagent recommendation, reaction type prediction, and yield prediction. \ModelName{} demonstrates comprehensive support across all tasks.
}
\label{tab:compare_downstream}
\end{table*}

\subsection{Downstream Tasks Dataset Construction}
\label{app:data_downstream}

In this section, we provide details on the data collection process for all downstream tasks of \PipeName{} introduced in Section \ref{sec:downstream_tasks}. Additionally, Table \ref{tab:compare_downstream} provides a comprehensive comparison of the capabilities of each method across these tasks.

\paragraph{Reaction Prediction.} We use USPTO-500-MT \citep{T5Chem, Mol-Instruction} and USPTO-full \citep{USPTO_patent, LLaSMol} datasets for reaction prediction. The training set of \citet{Mol-Instruction} has been chosen for its wide usage \citep{BioT5, BioT5+, nach0, InstructMol, ChemDFM}. However, while several previous works have reported near-optimal accuracy on the test set of \citet{Mol-Instruction}, we argue that most models still fail in real-world hard cases. To enhance the original test set's complexity, we add more challenging cases from \citet{LLaSMol}'s test set based on Bemis-Murcko scaffolds \citep{Bemis-Murcko}. This ensures lower similarity between train and test sets. The new test set has 1,000 samples to thoroughly evaluate the model's generalization ability. 

\noindent
\paragraph{Reaction Condition Prediction.} The reaction condition prediction tasks use combined data from TextReact \citep{TextReact} and Mol-Instruction \citep{Mol-Instruction}, both sourced from the USPTO dataset. Following \citet{TextReact}, we further annotate reaction condition prediction into subtasks with reagents, catalysts, and solvents. Notably, 65.75\% of the training reactions and 68.47\% of the test reactions in \citet{TextReact} overlap with \citet{Mol-Instruction}. To ensure fair comparison and utilize the additional data, we create a new dataset by combining the overlapping reactions. The data is split into train/valid/test sets with a ratio of 8:1:1 for each task.

\noindent
\paragraph{Reagent Selection.} Our study utilizes the reagent selection dataset from ChemLLMBench \citep{ChemLLMBench}, comprising 4,255 valid samples originally sourced from the Suzuki High-Throughput Experimentation (HTE) dataset \citep{HTEData}. Each sample includes reactants, a product, and a list of candidate reagents. The objective is to select the most suitable reagent from the candidate list to facilitate the reaction. The dataset is divided into 3,955 training samples and 300 testing samples, maintaining the same test split as \citet{ChemLLMBench}.

% \daniel{This dataset evaluates the Suzuki coupling of 5 electrophiles and 7 nucleophiles across a matrix of 11 ligands (with one blank), 7 bases (with one blank), and 4 solvents, resulting in a total of 5,760 data points. (from \citet{ChemLLMBench}}

\noindent
\paragraph{Reaction Type Classification.}
For reaction type classification, we use the USPTO 1K TPL dataset \citep{RXNFP} derived from the USPTO patent database \citep{USPTO_patent}, which contains 445,115 reactions labeled with 1000 reaction classes. Keeping the original configuration, the dataset is split into 360,545 samples for training, 40,059 for validation, and 44,511 for testing.

\noindent
\paragraph{Yield Regression.}
In this task, we use the Buchwald-Hartwig dataset \citep{DFT} and the Suzuki-Miyaura dataset \citep{HTEData} collected from \citet{YieldBERT}. The Buchwald-Hartwig dataset contains 3,955 reactions, while the Suzuki-Miyaura dataset contains 5,760 reactions. We follow the approach of ChemLLMBench, using their predefined test sets (100 tests each). Notably, we convert it into a regression task, and the yield values are normalized to the range [0, 1].

\subsection{Discussion on License.} 
As depicted in Table~\ref{tab:license}, we elaborate on the origins and legal permissions associated with each data component utilized in the development of the \PipeName{}. This encompasses both biomolecular data and textual descriptions. Thorough scrutiny was conducted on all data origins to confirm compatibility with our research objectives and subsequent utilization. Proper and accurate citation of these data sources is consistently maintained throughout the paper.

\begin{table*}[ht]
    \centering
    \vskip 0.1in
    \scalebox{0.69}{
    \begin{tabular}{p{4cm}p{7cm}p{11cm}}
    \toprule
    \textsc{\textbf{Data Sources}} & \textsc{\textbf{License URL}} & \textsc{\textbf{License Note}} \\
    \midrule
    PubChem & \url{https://www.nlm.nih.gov/web_policies.html} & Works produced by the U.S. government are not subject to copyright protection in the United States. Any such works found on National Library of Medicine (NLM) Web sites may be freely used or reproduced without permission in the U.S. \\
    ChEBI & \url{https://creativecommons.org/licenses/by/4.0/} & You are free to: Share — copy and redistribute the material in any medium or format. Adapt — remix, transform, and build upon the material for any purpose, even commercially. \\
    IUPAC & \url{https://iupac.org/wp-content/uploads/2021/06/iupac-inchi-license_2020.pdf} & An "IUPAC license" generally refers to the permissions, guidelines, or rights associated with using the standards, software, data, or publications provided by the International Union of Pure and Applied Chemistry (IUPAC). This can include adhering to IUPAC’s chemical nomenclature guidelines in scientific communication, using their proprietary software or databases under specific licensing terms, and obtaining permissions to reproduce or adapt copyrighted materials.\\
    USPTO & \url{https://www.uspto.gov/learning-and-resources/open-data-and-mobility} & It can be freely used, reused, and redistributed by anyone. \\
    \bottomrule
    \end{tabular}
    }
    \caption{
    \footnotesize{
    Data resources and licenses utilized in data collection for \PipeName{}}.
    }
    \vskip -0.1in
    \label{tab:license}
\end{table*}

%% file: sec/appendix/b-details.tex
\section{Implementation Details}
\label{app:implement_details}

\subsection{Evaluation Metrics}
\label{app:evaluation_metrics}

We utilize a variety of metrics to comprehensively evaluate the performance of the models across different types of tasks. The key metrics used for each type of task are as follows.

\paragraph{Classification Tasks.} For classification tasks, we report the following metrics:

\begin{itemize}[itemsep=1pt,parsep=0pt,left=0pt]
  \item \textbf{Accuracy}: The ratio of correctly classified samples.
  % \item \textbf{Macro F1}: The unweighted mean of F1 scores calculated for each class. It treats all classes equally regardless of their sample sizes.
  % \item \textbf{Micro F1}: The F1 score is calculated globally by counting all classes' total true positives, false negatives, and false positives. It takes class imbalance into account.
  \item \textbf{CEN} \citep{CEN}: The CEN score is a measure of the overall entropy of a confusion matrix, which is used to evaluate classifiers in multi-class problems.
  \item \textbf{MCC} \citep{MCC}: The MCC score is a balanced measure of binary classification quality, considering true and false positives and negatives.
\end{itemize}

\paragraph{Regression Tasks.} For regression tasks, we consider the following metrics:

\begin{itemize}[itemsep=1pt,parsep=0pt,left=0pt]
  \item \textbf{MAE}: Mean Absolute Error, the average absolute difference between predicted and actual values.  
  \item \textbf{MSE}: Mean Squared Error, the average squared difference between predicted and actual values.
  \item \textbf{$R^2$}: The coefficient of determination, indicating the proportion of variance in the target variable that is predictable from the input features. 
\end{itemize}

\paragraph{Molecule Generation Tasks.} For tasks involving SMILES \citep{SMILES} representations of molecules, we calculate:

\begin{itemize}[itemsep=1pt,parsep=0pt,left=0pt]
  \item \textbf{Exact Match}: The proportion of predicted SMILES strings that exactly match the ground truth after canonicalization.
  \item \textbf{BLEU} \citep{BLEU}: The BLEU score treats the SMILES strings as text, measuring n-gram overlap between predictions and references.
  \item \textbf{Levenshtein Distance} \cite{Levenshtein}: The minimum number of single-character edits required to change the predicted SMILES into the reference.
  \item \textbf{RDKit Similarity} \citep{RDKit}: The Tanimoto similarity between RDKit fingerprints of the predicted and reference molecules.
  \item \textbf{MACCS Keys Similarity} \citep{MACCS}: The Tanimoto similarity between MACCS keys fingerprints of the molecules.
  \item \textbf{Morgan Fingerprint Similarity} \citep{Morgan}: The Tanimoto similarity between Morgan circular fingerprints of the molecules.
  \item \textbf{Validity}: The proportion of predicted SMILES strings that can be successfully parsed into valid molecule structures by RDKit.
\end{itemize}

Note that if the origin model is trained on SELFIES \citep{SELFIES}, we use \citet{SELFIESPackage} to translate the generated SELFIES to SMILES before evaluation.

\subsection{Experimental Details}
Here we detail the hyperparameters for \PipeName{} pretraining and SFT.

\paragraph{PRESTO Alignment Stage.} We employed the PubChem molecule caption dataset, comprising approximately 327K samples, for training over 5 epochs. Training was conducted using 8$\times$A6000 GPUs, with a total batch size of 128. AdamW optimizer was utilized with $\beta = (0.9, 0.999)$ and a learning rate of 2e-3, without weight decay. The learning rate was initially warmed up over 3\% of the total training steps, followed by a cosine decay schedule. The model's maximum sequence length was set to 2048 for the base LLM. To conserve CUDA memory, we employed DeepSpeed ZeRO-2 strategy and gradient checkpointing.

\noindent
\paragraph{PRESTO Domain Incremental Pretrain Stage.} Using the projector checkpoint from the alignment stage, training followed the fundamental settings of the alignment stage, with adjustments made to the total batch size, set to 64, and the learning rate, set to 2e-5. Due to the prohibitive costs associated with fully finetuning the base 7B LLMs and the extensive pretraining dataset, all experiments were limited to one epoch.

\noindent
\paragraph{Supervised Finetuning.} We utilize the updated projector and LLM weights from the pretraining stage and combine all downstream task training sets for joint model training. For the full finetuning experiment, we train for three epochs by default, using the same hyperparameters as in the pretraining stage except for setting the total batch size to 128. For the LoRA ablation, we set the peak learning rate to 8e-5.

% \subsection{Baselines}
% We briefly introduce the baselines:
% \begin{itemize}[itemsep=1pt,parsep=0pt,left=0pt]
% \item \textbf{Galactica}~\cite{Galactica}. 

% \item \textbf{Mol-Instruction}~\cite{Mol-Instruction}. 

% \item \textbf{MoleculeTransformers}~\cite{MolecularTransformer}. 

% \item \textbf{Chemformer}~\cite{Chemformer}. 

% \item \textbf{Retroformer}~\cite{Retroformer}.

% \item \textbf{TextChemT5}~\cite{Text+ChemT5}.

% \item \textbf{nach0}~\cite{nach0}.

% \item \textbf{LIaSMol}~\cite{LLaSMol}.

% \item \textbf{T5Chem}~\cite{T5Chem}.

% \item \textbf{ChemDFM}~\cite{ChemDFM}.

% \end{itemize}

%% file: sec/appendix/c-ablation-SFT.tex
\section{More Ablations}
\label{app:more_ablation}

This section extends Section \ref{sec:ablation_main} to introduce more findings according to the ablation experiments.

\subsection{Analyzing SFT}
\label{app:analyze_sft}

Here, we explore important aspects of supervised finetuning, such as parameters, training time, and data scaling.

\begin{figure*}[!hbt]
    \centering
    \begin{subfigure}{0.48\textwidth}
        \includegraphics[width=\textwidth]{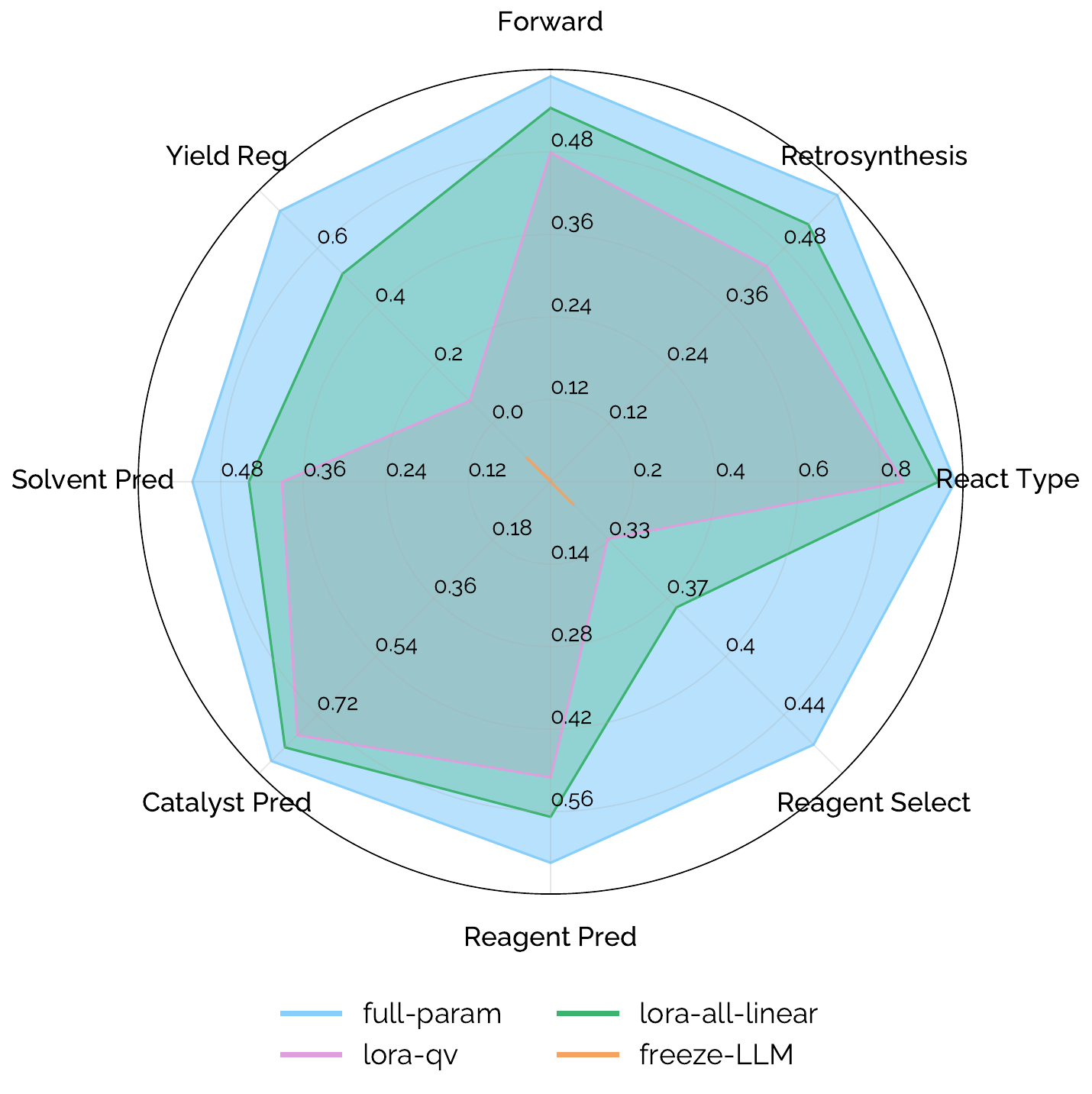}
        \caption{\textbf{\# Trainable Param Ablation}}
        \label{fig:sft_param}
    \end{subfigure}
    \hfill
    \begin{subfigure}{0.48\textwidth}
        \includegraphics[width=\textwidth]{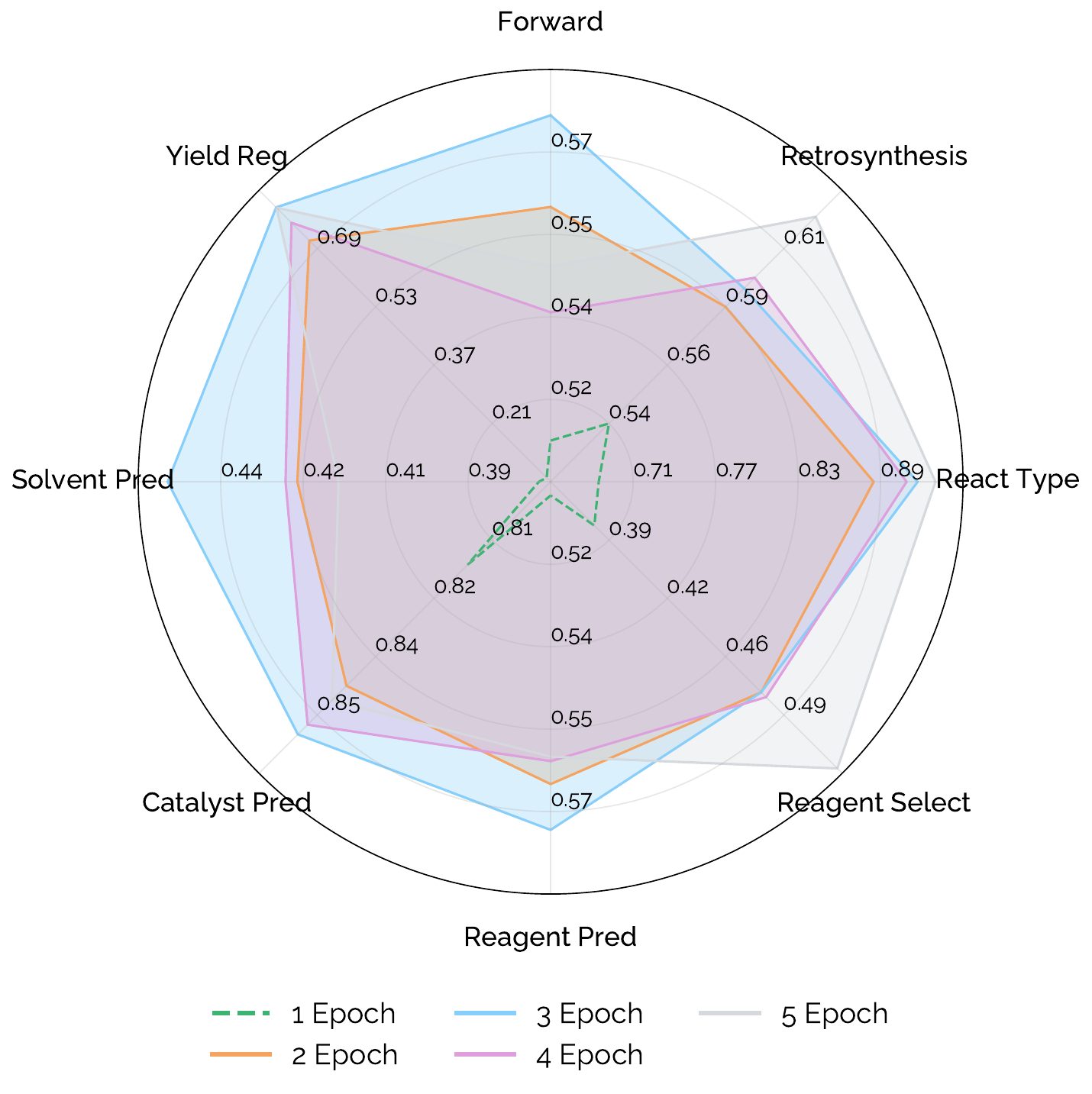}
        \caption{\textbf{Scaling SFT Train Time}}
        \label{fig:sft_train_time}
    \end{subfigure}
    \caption{\textbf{Performance analysis of different training strategies and dataset configurations.} (a) Ablation study on the trainable parameters in the LLM during SFT. An increase in trainable parameters consistently enhances performance. (b) Analysis of training duration impacts on SFT. Performance improves up to three epochs, while training for four epochs results in overfitting.}
\end{figure*}

\paragraph{Finding 5: Updating LLMs is essential.} We conducted an ablation study on the trainable parameters of LLMs during the SFT stage (Figure \ref{fig:sft_param}), progressing from not updating any LLM parameters to updating the attention block’s \texttt{q\_proj} and \texttt{v\_proj} layers with LoRA, then updating all linear layers except the \texttt{lm\_head} layer with LoRA, and finally fully finetuning all parameters. All experiments involved training for 3 epochs on the SFT dataset. We found that not updating the LLM parameters during SFT led to nearly zero performance, highlighting the necessity of parameter updates for adapting to downstream tasks. Incorporating LoRA modules significantly boosted performance, and adding more trainable LoRA modules consistently improved results. Moreover, when computational resources allow, full-tuning outperforms LoRA-tuning across various downstream tasks.
% \begin{figure}[!hbt]
%     \centering
%     \includegraphics[width=0.5\textwidth]{figs/sft_llm_param_ablation_radar.pdf}
%     \caption{\footnotesize{Ablation study on the trainable LLM parameters during SFT. With more trainable parameters, the performance grows consistently.}}
%     \label{fig:sft_param}
% \end{figure}

\paragraph{Finding 6: Balancing SFT training time optimizes downstream task performance.}
We investigate the impact of SFT training time on a subset of our SFT training dataset (1/7 size, detailed in the Appendix). Unlike existing Vision LMs, which typically undergo only one epoch of training, we compare performance across different numbers of epochs. We observe severe underfitting with only one epoch of training. Surprisingly, we find steady improvement across all tasks when trained for up to three epochs but encounter overfitting when training to four epochs, leading to performance degradation. In conclusion, we recommend training for three epochs for optimal performance on downstream tasks.
% as training less causes underfitting, and longer durations lead to overfitting.
% fix FLOPS(1epoch), reduce SFT dataset to 1/2, 1/4, 1/8, only for FS & RS; compared on Random-Split-Test and Scaffold-Split-Test
% backup: direct compare subset vs full

% \begin{figure}[!hbt]
%     \centering
%     \includegraphics[width=0.5\textwidth]{figs/scaling_sft_train_time.pdf}
%     \caption{\footnotesize{Scaling on the SFT train time.}}
%     \label{fig:sft_train_time}
% \end{figure}

\begin{figure*}[!hbt]
    \centering
    \includegraphics[width=0.8\textwidth]{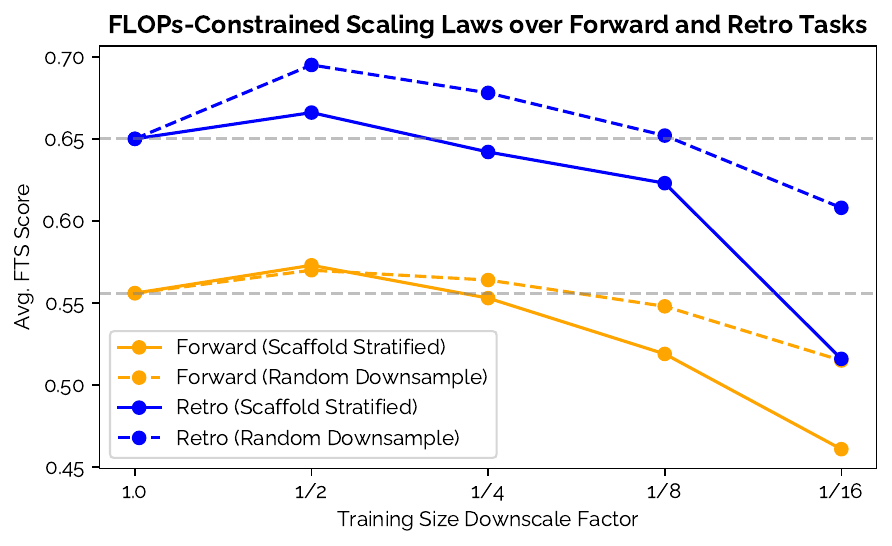}
    \caption{\textbf{Impact of SFT training dataset coverage and diversity on downstream task performance.} Training up to four epochs with repeated data resulted in negligible changes in loss compared to using unique data. Maintaining the number of scaffold clusters even when the training set size was halved led to higher performance on the test set.}
    \label{fig:sft_train_data}
\end{figure*}

\paragraph{Finding 7: Coverage and diversity of SFT dataset are critical for better results.} We examined the impact of data repetition (i.e., allocating FLOPs across multiple epochs on the same data) and SFT-data size on downstream tasks. In our experiments on forward and retrosynthesis prediction, we fixed the training FLOPs (equivalent to the FLOPs used to train for 1 epoch with the full dataset) and successively halved the training dataset while doubling the number of training epochs. We used two subsampling methods: (1) random subsampling and (2) hierarchical subsampling based on scaffold clustering. Figure \ref{fig:sft_train_data} revealed that for a fixed compute budget, training up to four epochs with repeated data resulted in negligible changes in loss compared to using unique data. Moreover, we found that the coverage and diversity of the SFT training set are crucial; even when the training set size was halved, maintaining the number of scaffold clusters led to higher performance on the test set.

%% file: sec/appendix/d-template.tex
\section{Instruction Templates}
\label{app:template_main}
In this section, we provide a basic description of the instruction templates utilized in \PipeName{}. These templates are designed to guide the model during pretraining and downstream tasks. We have a variety of templates for each task, and we present a randomly selected template in this part.

\subsection{Template for Pretraining}
\label{app:template_pretrain}
Here are six templates used in the pretraining stage of \PipeName{}:

\begin{enumerate}[itemsep=1pt,parsep=0pt,topsep=0pt]
    \item PubChem Caption (Table \ref{table:caption_generation_template})
    \item IUPAC to Formula (Table \ref{table:iupac_to_formula_template})
    \item IUPAC to SMILES (Table \ref{table:iupac_to_smiles_template})
    \item Molecule Graph to Formula (Table \ref{table:representation_to_formula_template})
\item Molecule Graph to IUPAC (Table \ref{table:representation_to_iupac_template})
\item Molecule Graph to SMILES (Table \ref{table:representation_to_smiles_template})
\end{enumerate}

\input{sec/template/template_caption}
\input{sec/template/template_i2f}
\input{sec/template/template_i2s}
\input{sec/template/template_g2f}
\input{sec/template/template_g2i}
\input{sec/template/template_g2s}

\subsection{Template for Downstream Tasks}
\label{app:template_downstream}
Here are 10 templates used for downstream tasks of \PipeName{}:
\begin{enumerate}[itemsep=1pt,parsep=0pt,topsep=0pt]
\item Forward Prediction (Table \ref{table:forward_prediction_template})
\item Retrosynthesis Prediction (Table \ref{table:retro_prediction_template})
\item Catalyst Prediction (Table \ref{table:catalyst_prediction_template})
\item Reagent Prediction (Table \ref{table:reagent_prediction_template})
\item Solvent Prediction (Table \ref{table:solvent_prediction_template})
\item Reagent Selection (Table \ref{table:reagent_selection_template})
\item Ligand Selection (Table \ref{table:ligand_selection_template})
\item Solvent Selection (Table \ref{table:solvent_selection_template})
\item Yield Prediction (Table \ref{table:yield_prediction_template})
\item Reaction Type Classification (Table \ref{table:reaction_class_prediction_template})
\end{enumerate}

\input{sec/template/template_forward}
\input{sec/template/template_retro}
\input{sec/template/template_catalyst_pred}
\input{sec/template/template_reagent_pred}
\input{sec/template/template_solvent_pred}
\input{sec/template/template_reagent_select}
\input{sec/template/template_ligand_select}
\input{sec/template/template_solvent_select}
\input{sec/template/template_yield}
\input{sec/template/template_reaction_tpl}

%% file: sec/template/template_caption.tex
\begin{table*}[!hbt]
\centering
\begin{tabular}{p{0.98\linewidth}}
\toprule
\rowcolor{gray!20}
\textbf{System Prompt:}\\
\hline

You are a chemist. Now you are given a representation of a molecule. Please help me to understand the molecule.\\

\hline
\rowcolor{gray!20}
\textbf{Instruction:}\\
\hline

Provide a brief overview of this molecule: \textcolor{orange}{<molecule>}.\\

\hline
\rowcolor{gray!20}  
\textbf{Output:}\\
\hline
Sure! Here is a description of this molecule.
\textcolor{teal}{<caption>}.\\
\bottomrule
\end{tabular}
\caption{\textbf{An instruction template for generating captions for molecules.} This template guides the generation of a brief description or overview of the given molecule representation.}
\label{table:caption_generation_template}
\end{table*}

%% file: sec/template/template_i2f.tex
\begin{table*}[!hbt]
\centering
\begin{tabular}{p{0.98\linewidth}}
\toprule
\rowcolor{gray!20}
\textbf{System Prompt:}\\
\hline

You are a chemist. Please follow the instructions to convert the structure to the corresponding name.\\

\hline
\rowcolor{gray!20}
\textbf{Instruction:}\\
\hline

\textcolor{orange}{<input>} is the IUPAC name of a molecule. Please give its molecular formula.\\

\hline
\rowcolor{gray!20}  
\textbf{Output:}\\
\hline

The molecular formula is \textcolor{teal}{<output>}.\\
\bottomrule
\end{tabular}
\caption{\textbf{An instruction template for converting IUPAC names to molecular formulas.} This template guides the conversion from the given IUPAC name to its corresponding molecular formula.}
\label{table:iupac_to_formula_template}
\end{table*}

%% file: sec/template/template_i2s.tex
\begin{table*}[!hbt]
\centering
\begin{tabular}{p{0.98\linewidth}}
\toprule
\rowcolor{gray!20}
\textbf{System Prompt:}\\
\hline

You are a chemist. Please follow the instructions to convert the structure to the corresponding name.\\

\hline
\rowcolor{gray!20}
\textbf{Instruction:}\\
\hline

Convert the IUPAC name of a molecule \textcolor{orange}{<input>} into SMILES representation.\\

\hline
\rowcolor{gray!20}  
\textbf{Output:}\\
\hline

The SMILES representation is \textcolor{teal}{<output>}.\\
\bottomrule
\end{tabular}
\caption{\textbf{An instruction template for converting IUPAC names to SMILES representations.} This template guides the conversion from the given IUPAC name to its corresponding SMILES representation.}
\label{table:iupac_to_smiles_template}
\end{table*}

%% file: sec/template/template_g2f.tex
\begin{table*}[!hbt]
\centering
\begin{tabular}{p{0.98\linewidth}}
\toprule
\rowcolor{gray!20}
\textbf{System Prompt:}\\
\hline

You are a chemist. Please follow the instructions to convert the structure to the corresponding name.\\

\hline
\rowcolor{gray!20}
\textbf{Instruction:}\\
\hline

\textcolor{orange}{<input>} is the representation of a molecule. What is its molecular formula?\\

\hline
\rowcolor{gray!20}  
\textbf{Output:}\\
\hline

The molecular formula is \textcolor{teal}{<output>}.\\
\bottomrule
\end{tabular}
\caption{\textbf{An instruction template for converting molecular graph to molecular formula.} This template guides the conversion from the given graph representation to its corresponding molecular formula.}
\label{table:representation_to_formula_template}
\end{table*}

%% file: sec/template/template_g2i.tex
\begin{table*}[!hbt]
\centering
\begin{tabular}{p{0.98\linewidth}}
\toprule
\rowcolor{gray!20}
\textbf{System Prompt:}\\
\hline

You are a chemist. Please follow the instructions to convert the structure to the corresponding name.\\

\hline
\rowcolor{gray!20}
\textbf{Instruction:}\\
\hline

\textcolor{orange}{<input>} is the representation of a molecule. What is its IUPAC name?\\

\hline
\rowcolor{gray!20}  
\textbf{Output:}\\
\hline

The IUPAC name is \textcolor{teal}{<output>}.\\
\bottomrule
\end{tabular}
\caption{\textbf{An instruction template for converting molecule graph to IUPAC name.} This template guides the conversion from the given graph representation to its corresponding IUPAC name.}
\label{table:representation_to_iupac_template}
\end{table*}

%% file: sec/template/template_g2s.tex
\begin{table*}[!hbt]
\centering
\begin{tabular}{p{0.98\linewidth}}
\toprule
\rowcolor{gray!20}
\textbf{System Prompt:}\\
\hline

You are a chemist. Please follow the instructions to convert the structure to the corresponding name.\\

\hline
\rowcolor{gray!20}
\textbf{Instruction:}\\
\hline

The representation of a certain molecule is \textcolor{orange}{<input>}. Can you provide its SMILES representation?\\

\hline
\rowcolor{gray!20}  
\textbf{Output:}\\
\hline

The SMILES representation is \textcolor{teal}{<output>}.\\
\bottomrule
\end{tabular}
\caption{\textbf{An instruction template for converting the molecule graph to SMILES representation.} This template guides the conversion from the given graph representation to its corresponding SMILES representation.}
\label{table:representation_to_smiles_template}
\end{table*}

%% file: sec/template/template_forward.tex
\begin{table*}[!hbt]
\centering
\begin{tabular}{p{0.98\linewidth}}
\toprule
\rowcolor{gray!20}
\textbf{System:}\\
\hline

You are a chemist. Your task is to predict the SMILES representation of the product molecule, given the molecule representations of the reactants.\\

\hline
\rowcolor{gray!20}
\textbf{Instruction:}\\
\hline

Using \textcolor{orange}{<reactant\_1>}.\textcolor{red}{<reactant\_2>}.\textcolor{purple}{<reactant\_3>} as the reactants and reagents, tell me the potential product.\\

\hline
\rowcolor{gray!20}  
\textbf{Output:}\\
\hline

Sure. A potential product: \textcolor{blue}{<product\_1>}.\textcolor{cyan}{<product\_2>}.\\
\bottomrule
\end{tabular}
\caption{\textbf{An instruction template for forward prediction.} This template guides the prediction of the product based on the given reactants and reagents. The reactants and reagents are specified, and the model must predict the potential product from the reaction.}
\label{table:forward_prediction_template}
\end{table*}

%% file: sec/template/template_retro.tex
\begin{table*}[!hbt]
\centering
\begin{tabular}{p{0.98\linewidth}}
\toprule
\rowcolor{gray!20}
\textbf{System:}\\
\hline

You are a chemist. Your task is to predict the SMILES representation of the reactant molecules, given the molecule representations of the product.\\

\hline
\rowcolor{gray!20}
\textbf{Instruction:}\\
\hline

Using \textcolor{orange}{<product\_1>}.\textcolor{red}{<product\_2>}.\textcolor{purple}{<product\_3>} as the products, predict the possible reactants that could have been utilized to synthesize these products.\\

\hline
\rowcolor{gray!20}  
\textbf{Output:}\\
\hline

Here are possible reactants: \textcolor{blue}{<reactant\_1>}.\textcolor{cyan}{<reactant\_2>}.\\
\bottomrule
\end{tabular}
\caption{\textbf{An instruction template for retrosynthesis prediction.} This template guides the prediction of the possible reactants based on the given product. The product is specified, and the model must predict the reactants that could have been used to synthesize this product.}
\label{table:retro_prediction_template}
\end{table*}

%% file: sec/template/template_catalyst_pred.tex
\begin{table*}[!hbt]
\centering
\begin{tabular}{p{0.98\linewidth}}
\toprule
\rowcolor{gray!20}
\textbf{System Prompt:}\\
\hline

You are a chemist. Now, you are given a reaction equation. Your task is to predict the SMILES representation of the catalyst, given molecule representation of the reaction.\\

\hline
\rowcolor{gray!20}
\textbf{Instruction:}\\
\hline

Based on the given chemical reaction: \textcolor{orange}{<reactant\_1>}.\textcolor{red}{<reactant\_2>}.\textcolor{purple}{<reactant\_3>} >> \textcolor{blue}{<product\_1>}.\textcolor{violet}{<product\_2>}, propose some likely catalysts that might have been utilized.\\

\hline
\rowcolor{gray!20}  
\textbf{Output:}\\
\hline

A possible catalyst can be \textcolor{teal}{<catalyst>}.\\
\bottomrule
\end{tabular}
\caption{\textbf{An instruction template for catalyst prediction.} This template guides the prediction of possible catalysts based on the given reaction components. The reactants and products are specified, and the model must predict the potential catalyst from the reaction.}
\label{table:catalyst_prediction_template}
\end{table*}

%% file: sec/template/template_reagent_pred.tex
\begin{table*}[!hbt]
\centering
\begin{tabular}{p{0.98\linewidth}}
\toprule
\rowcolor{gray!20}
\textbf{System Prompt:}\\
\hline

You are a chemist. Now, you are given a reaction equation. Your task is to predict the SMILES representation of the reagents, given molecule representation of the reaction.\\

\hline
\rowcolor{gray!20}
\textbf{Instruction:}\\
\hline

Based on the given chemical reaction: \textcolor{orange}{<reactant\_1>}.\textcolor{red}{<reactant\_2>}.\textcolor{purple}{<reactant\_3>} >> \textcolor{blue}{<product\_1>}.\textcolor{violet}{<product\_2>}, propose some likely reagents that might have been utilized.\\

\hline
\rowcolor{gray!20}  
\textbf{Output:}\\
\hline

A possible reagent can be \textcolor{teal}{<reagent>}.\\
\bottomrule
\end{tabular}
\caption{\textbf{An instruction template for reagent prediction.} This template guides the prediction of possible reagents based on the given reaction components. The reactants and products are specified, and the model must predict the potential reagent from the reaction.}
\label{table:reagent_prediction_template}
\end{table*}

%% file: sec/template/template_solvent_pred.tex
\begin{table*}[!hbt]
\centering
\begin{tabular}{p{0.98\linewidth}}
\toprule
\rowcolor{gray!20}
\textbf{System Prompt:}\\
\hline

You are a chemist. Now, you are given a reaction equation. Your task is to predict the SMILES representation of the solvents, given molecule representation of the reaction.\\

\hline
\rowcolor{gray!20}
\textbf{Instruction:}\\
\hline

Based on the given chemical reaction: \textcolor{orange}{<reactant\_1>}.\textcolor{red}{<reactant\_2>}.\textcolor{purple}{<reactant\_3>} >> \textcolor{blue}{<product\_1>}.\textcolor{violet}{<product\_2>}, propose some likely solvents that might have been utilized.\\

\hline
\rowcolor{gray!20}  
\textbf{Output:}\\
\hline

A possible solvent can be \textcolor{teal}{<solvent>}.\\
\bottomrule
\end{tabular}
\caption{\textbf{An instruction template for solvent prediction.} This template guides the prediction of possible solvents based on the given reaction components. The reactants and products are specified, and the model must predict the potential solvent from the reaction.}
\label{table:solvent_prediction_template}
\end{table*}

%% file: sec/template/template_reagent_select.tex
\begin{table*}[!hbt]
\centering
\begin{tabular}{p{0.98\linewidth}}
\toprule
\rowcolor{gray!20}
\textbf{System Prompt:}\\
\hline

You are an expert chemist. Given one reactant, two reagents, and one solvent of a Suzuki reaction, predict the optimal reactant that maximizes the yield with the rest of the reaction components. Only return the option from the given list.\\

\hline
\rowcolor{gray!20}
\textbf{Instruction:}\\
\hline

Given the rest of the reaction components: \textcolor{orange}{<reactant\_1>} > \textcolor{purple}{<reagent\_1>}.\textcolor{purple}{<reagent\_2>} >> \textcolor{blue}{<solvent>}. \\
Select the optimal reactant: \textcolor{teal}{<reactant\_2>}.\textcolor{violet}{<reactant\_3>} \\

\hline
\rowcolor{gray!20}  
\textbf{Output:}\\
\hline

Optimal reactant: \textcolor{violet}{<reactant\_3>}.\\
\bottomrule
\end{tabular}
\caption{\textbf{An instruction template for reagent selection.} This template guides the prediction of the optimal reactant based on the given reaction components. The reactant, reagents, and solvent are specified, and the model must choose the best reactant from the provided list.}
\label{table:reagent_selection_template}
\end{table*}

%% file: sec/template/template_ligand_select.tex
\begin{table*}[!hbt]
\centering
\begin{tabular}{p{0.98\linewidth}}
\toprule
\rowcolor{gray!20}
\textbf{System Prompt:}\\
\hline

You are an expert chemist. Given two reactants, one reagent, and one solvent of a Suzuki reaction, predict the optimal ligand that maximizes the yield with the rest of the reaction components. Only return the option from the given list.\\

\hline
\rowcolor{gray!20}
\textbf{Instruction:}\\
\hline

Given the rest of the reaction components: \textcolor{orange}{<reactant\_1>}.\textcolor{red}{<reactant\_2>} >> \textcolor{purple}{<reagent>}.\textcolor{blue}{<solvent>}. \\
Select the optimal ligand: \textcolor{teal}{<ligand\_1>}.\textcolor{violet}{<ligand\_2>} \\

\hline
\rowcolor{gray!20}  
\textbf{Output:}\\
\hline

Optimal ligand: \textcolor{teal}{<ligand\_1>}.\\
\bottomrule
\end{tabular}
\caption{\textbf{An instruction template for ligand selection.} This template guides the prediction of the optimal ligand based on the given reaction components. The reactants, reagents, and solvents are specified, and the model must choose the best ligand from the provided list.}
\label{table:ligand_selection_template}
\end{table*}

%% file: sec/template/template_solvent_select.tex
\begin{table*}[!hbt]
\centering
\begin{tabular}{p{0.98\linewidth}}
\toprule
\rowcolor{gray!20}
\textbf{System Prompt:}\\
\hline

You are an expert chemist. Given two reactants, one ligand, and one base of a Suzuki reaction, predict the optimal solvent that maximizes the yield with the rest of the reaction components. Only return the option from the given list.\\

\hline
\rowcolor{gray!20}
\textbf{Instruction:}\\
\hline

Given the rest of the reaction components: \textcolor{orange}{<reactant\_1>}.\textcolor{red}{<reactant\_2>} >> \textcolor{purple}{<ligand>}.\textcolor{blue}{<base>}. \\
Select the optimal solvent: \textcolor{teal}{<solvent\_1>}.\textcolor{violet}{<solvent\_2>} \\

\hline
\rowcolor{gray!20}  
\textbf{Output:}\\
\hline

Optimal solvent: \textcolor{violet}{<solvent\_2>}.\\
\bottomrule
\end{tabular}
\caption{\textbf{An instruction template for solvent selection.} This template guides the prediction of the optimal solvent based on the given reaction components. The reactants, ligand, and base are specified, and the model must choose the best solvent from the provided list.}
\label{table:solvent_selection_template}
\end{table*}

%% file: sec/template/template_yield.tex
\begin{table*}[!hbt]
\centering
\begin{tabular}{p{0.98\linewidth}}
\toprule
\rowcolor{gray!20}
\textbf{System Prompt:}\\
\hline

You are a chemist. Now, you are given a reaction equation. Your task is to predict the yield ratio of the reaction. The return value should be in the range of 0-1. The higher the value, the more likely the reaction is to occur.\\

\hline
\rowcolor{gray!20}
\textbf{Instruction:}\\
\hline

Based on the given chemical reaction: \textcolor{orange}{<reactant\_1>}.\textcolor{red}{<reactant\_2>}.\textcolor{purple}{<reactant\_3>} >> \textcolor{blue}{<product\_1>}.\textcolor{violet}{<product\_2>}, what is the yield ratio of the reaction?\\

\hline
\rowcolor{gray!20}  
\textbf{Output:}\\
\hline

The yield ratio is \textcolor{teal}{<ratio>}.\\
\bottomrule
\end{tabular}
\caption{\textbf{An instruction template for yield prediction.} This template guides the prediction of the yield ratio based on the given reaction components. The reactants and products are specified, and the model must predict the yield ratio from the reaction.}
\label{table:yield_prediction_template}
\end{table*}

%% file: sec/template/template_reaction_tpl.tex
\begin{table*}[!hbt]
\centering
\begin{tabular}{p{0.98\linewidth}}
\toprule
\rowcolor{gray!20}
\textbf{System Prompt:}\\
\hline

You are a chemist. Now, you are given a reaction equation. Your task is to predict the class of the reaction. Your task is to predict the class number of the reaction.\\

\hline
\rowcolor{gray!20}
\textbf{Instruction:}\\
\hline

Based on the given chemical reaction: \textcolor{orange}{<reactant\_1>}.\textcolor{red}{<reactant\_2>}.\textcolor{purple}{<reactant\_3>} >> \textcolor{blue}{<product\_1>}.\textcolor{violet}{<product\_2>}, predict the class number of the reaction.\\

\hline
\rowcolor{gray!20}  
\textbf{Output:}\\
\hline

The class number is \textcolor{teal}{<class\_number>}.\\
\bottomrule
\end{tabular}
\caption{\textbf{An instruction template for reaction type classification.} This template guides the prediction of the reaction class number based on the given reaction components. The reactants and products are specified, and the model must predict the reaction class number from the reaction.}
\label{table:reaction_class_prediction_template}
\end{table*}

%% file: sec/appendix/e-examples.tex
\section{Case Studies}
\begin{figure*}[!ht]
    \centering
    \includegraphics[width=1\linewidth]{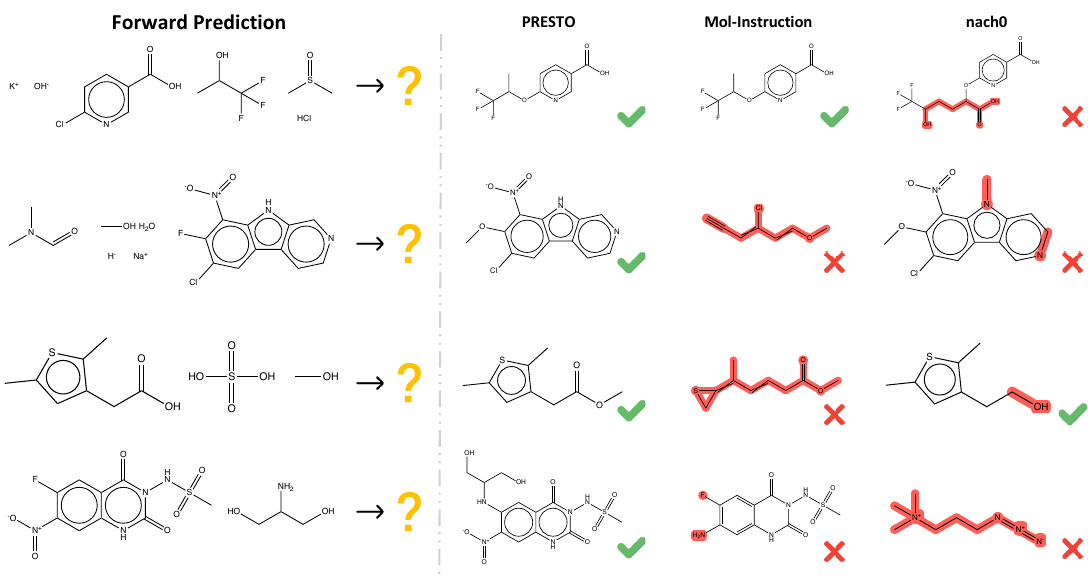}
    \caption{More examples of the \textbf{Forward Prediction} task. We include Mol-Instruction~\cite{Mol-Instruction} and nach0~\cite{nach0} as baselines.}
    \label{fig:appendix-forward}
\end{figure*}

\begin{figure*}[!ht]
    \centering
    \includegraphics[width=1\linewidth]{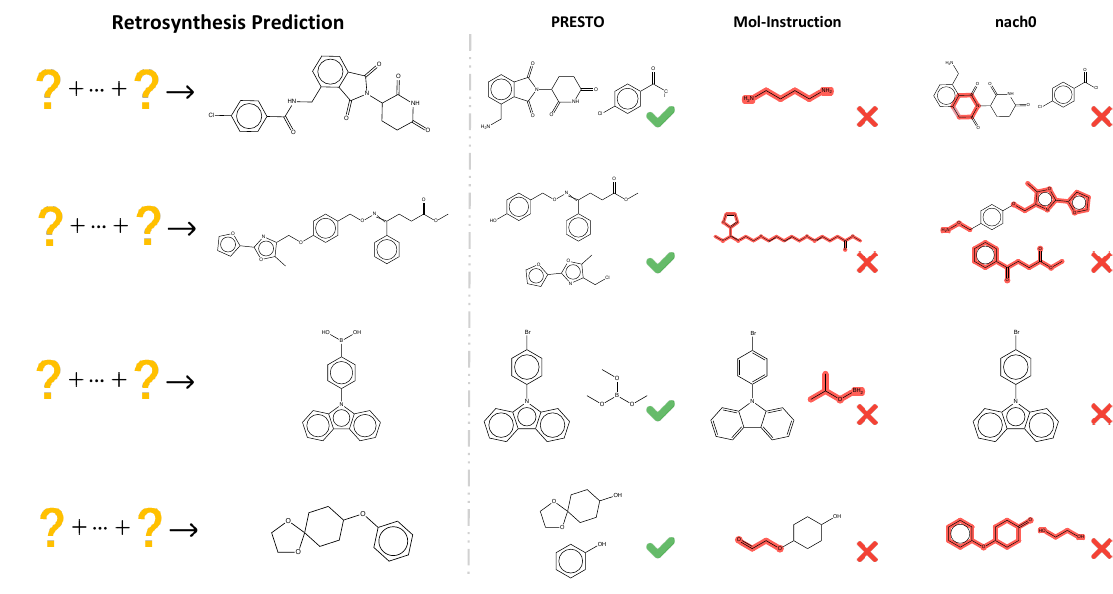}
    \caption{More examples of the \textbf{Retrosynthesis Prediction} task. We include Mol-Instruction~\cite{Mol-Instruction} and nach0~\cite{nach0} as baselines.}
    \label{fig:appendix-retro}
\end{figure*}

\begin{figure*}[!ht]
    \centering
    \includegraphics[width=1\linewidth]{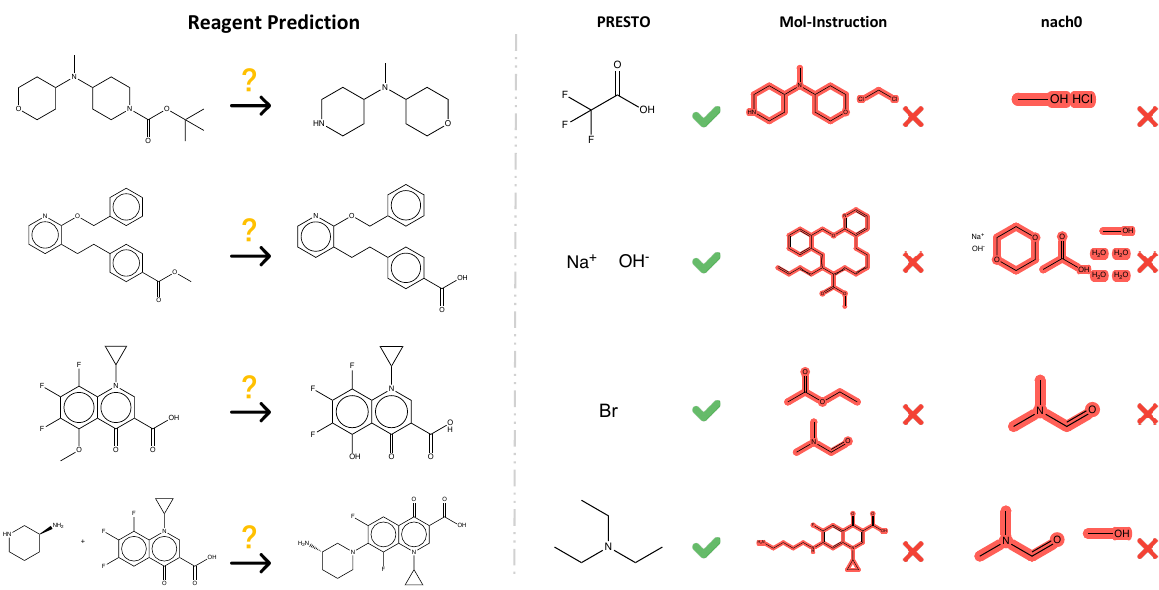}
    \caption{More examples of the \textbf{Reagent Prediction} task. We include Mol-Instruction~\cite{Mol-Instruction} and nach0~\cite{nach0} as baselines.}
    \label{fig:appendix-reagent-pred}
\end{figure*}

\begin{figure*}[!ht]
    \centering
    \includegraphics[width=1\linewidth]{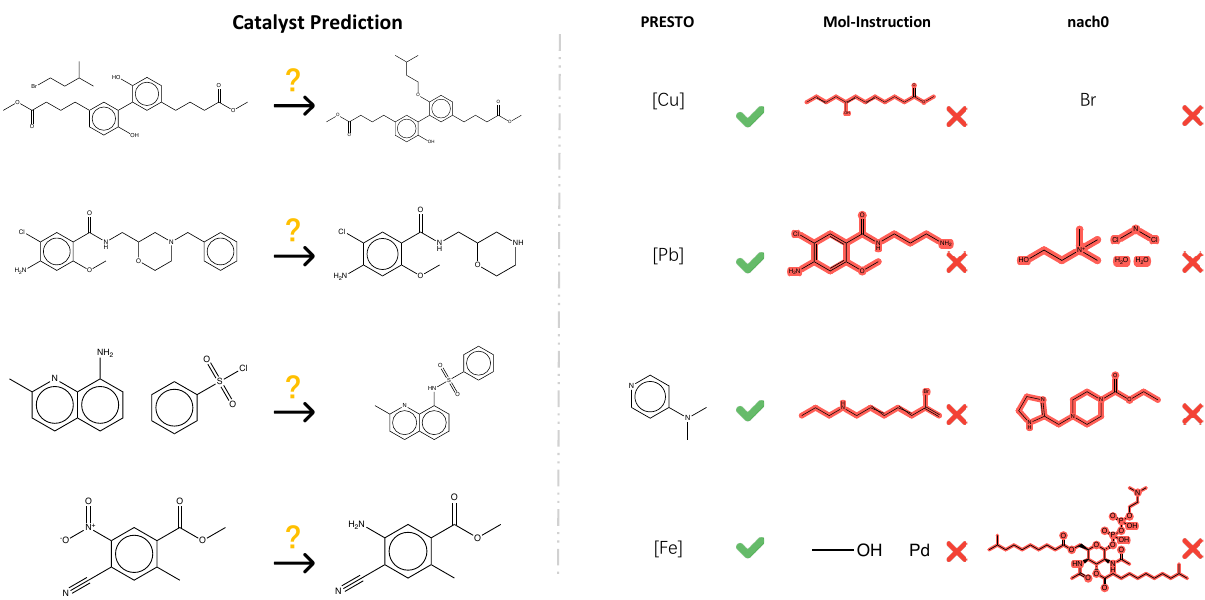}
    \caption{More examples of the \textbf{Catalyst Prediction} task. We include Mol-Instruction~\cite{Mol-Instruction} and nach0~\cite{nach0} as baselines.}
    \label{fig:appendix-catalyst-pred}
\end{figure*}

\begin{figure*}[!ht]
    \centering
    \includegraphics[width=1\linewidth]{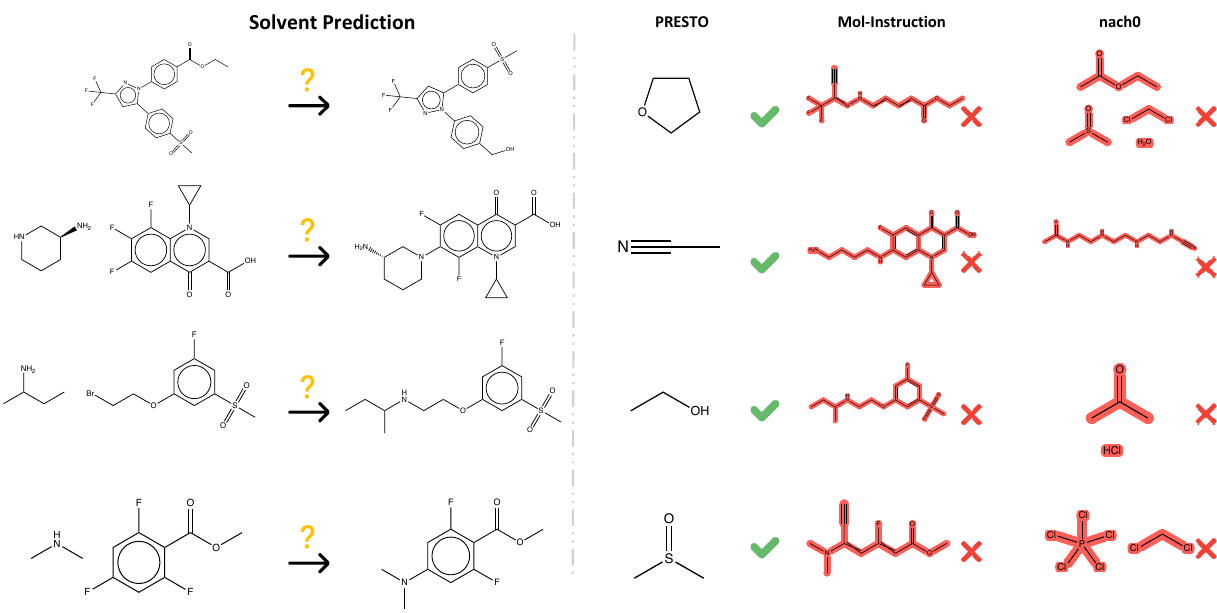}
    \caption{More examples of the \textbf{Solvent Prediction} task. We include Mol-Instruction~\cite{Mol-Instruction} and nach0~\cite{nach0} as baselines.}
    \label{fig:appendix-solvent-pred}
\end{figure*}

We show some selected cases for forward prediction (Table~\ref{fig:appendix-forward}), retrosynthesis prediction (Table~\ref{fig:appendix-retro}), reagent prediction (Table~\ref{fig:appendix-reagent-pred}), solvent prediction (Table~\ref{fig:appendix-solvent-pred}), and catalyst prediction tasks (Table~\ref{fig:appendix-catalyst-pred}).

